\documentclass[11pt]{article}

\usepackage[final]{acl}

\usepackage{times}
\usepackage{latexsym}
\usepackage{booktabs}
\usepackage{marvosym}

\usepackage[T1]{fontenc}

\usepackage[utf8]{inputenc}
\usepackage[normalem]{ulem}

\usepackage{microtype}

\usepackage{inconsolata}
\usepackage{url}

\usepackage{graphicx}
\usepackage{amsmath}
\usepackage{amssymb}
\usepackage{cuted}
\usepackage{capt-of}
\usepackage{pifont}
\usepackage{amssymb}
\usepackage{tabularx}
\usepackage{booktabs}

\usepackage{xcolor} 
\usepackage{colortbl}
\definecolor{cvprblue}{rgb}{0.21,0.49,0.74}  
\definecolor{mygreen}{RGB}{0,150,0}
\definecolor{myred}{RGB}{200,0,0}
\definecolor{lightblue}{RGB}{230, 242, 255}
\usepackage{multirow}
\usepackage{enumitem}

\usepackage[most]{tcolorbox}
\usepackage{xcolor}
\usepackage{listings}
\usepackage{caption}
\usepackage{hyperref}

\definecolor{promptblue}{RGB}{102,167,208}
\definecolor{promptgray}{gray}{0.45}

\title{TouchThinker: Scaling Tactile Commonsense Reasoning to the Open World with Large-scale Data and Action-aware Representation}

\author{
\textbf{Kailin Lyu\textsuperscript{1,2}},
\textbf{Di Wu\textsuperscript{1}},
\textbf{Pengwei Zhang\textsuperscript{1,2}},
\textbf{Yuhang Zheng\textsuperscript{3}},
\textbf{Yingxin Lai\textsuperscript{4}},
\\
\textbf{Long Xiao\textsuperscript{1}},
\textbf{Kangyi Wu\textsuperscript{5}},
\textbf{Pengna Li\textsuperscript{5}},
\textbf{Chen Gao\textsuperscript{3}},
\textbf{Lianyu Hu\textsuperscript{6}},
\\
\textbf{Xiaobin Hu\textsuperscript{3,$\dagger$}},
\textbf{Jie Hao\textsuperscript{1,$\dagger$}},
\textbf{Ce Hao\textsuperscript{2,$\dagger$}},
\textbf{Weihao Yuan\textsuperscript{7}},
\textbf{Shuicheng Yan\textsuperscript{3,$\dagger$}}
\\
\\
\textsuperscript{1}Institute of Automation, Chinese Academy of Sciences
\\
\textsuperscript{2}Zhongguancun Academy,
\textsuperscript{3}National University of Singapore,
\textsuperscript{4}Xiamen University
\\
\textsuperscript{5}Xi'an Jiaotong University,
\textsuperscript{6}Nanyang Technological University,
\textsuperscript{7}Nanjing University
\\
\textsuperscript{$\dagger$}Corresponding authors}

\begin{document}
\maketitle

\begin{strip}
\centering
\includegraphics[width=\textwidth]{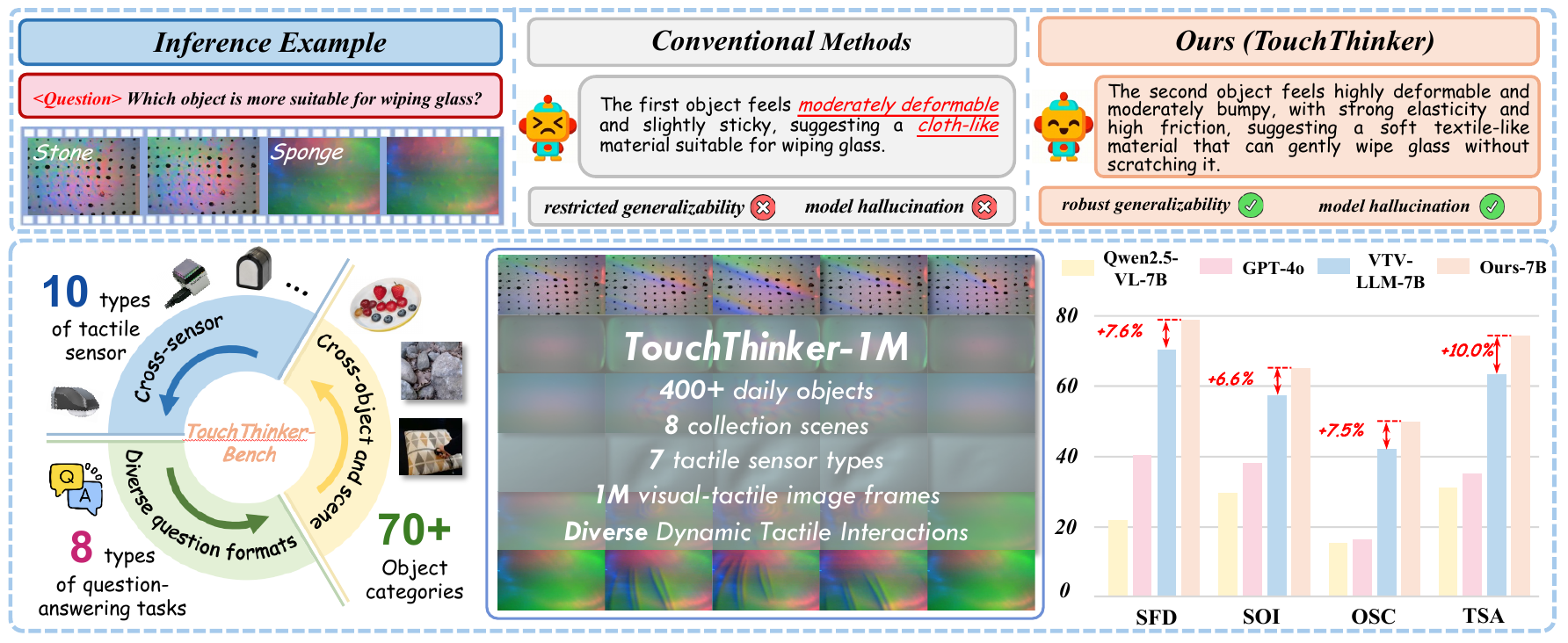}
\captionof{figure}{\textbf{The core contributions of TouchThinker.} (1) We propose an action-aware tactile encoder to enable efficient tactile representation learning. Upon it, we construct TouchThinker-1M, a million-scale, multi-source visuo-tactile dataset that expands the scope of tactile reasoning. (2) We further introduce TouchThinker-Bench to support multi-dimensional evaluation of open-world tactile reasoning. (3) Across multiple tasks, TouchThinker achieves significant improvements over prior methods.}
\label{intro}
\end{strip}

\begin{abstract}
Touch is a key modality for embodied agents to understand the physical world. Although recent work has incorporated tactile signals into language systems for tactile commonsense reasoning, scaling such systems to realistic open-world settings remains challenging due to two key bottlenecks: (1) current tactile reasoning datasets remain limited in format and scale, providing insufficient supervision for reasoning from tactile observations to physical commonsense and hindering the learning of transferable tactile commonsense; (2) tactile signals are inherently redundant and action-specific, yet existing methods often overlook these properties, resulting in inefficient representations with limited semantic expressiveness. To address these limitations, we propose TouchThinker, a tactile-language framework that scales tactile commonsense reasoning to the open world from both data and representation perspectives. First, we construct TouchThinker-1M, a million-scale, multi-source tactile reasoning dataset covering \textbf{415} objects, \textbf{8} scenarios, and \textbf{7} sensor types, providing a solid data foundation for open-world generalization. We further introduce TouchThinker-Bench, an open-world benchmark with more realistic and diverse tasks. Then, we propose an action-aware modeling mechanism to improve tactile representation efficiency and enable efficient reasoning. Experimental results demonstrate that TouchThinker achieves competitive performance against state-of-the-art models across multiple datasets. Our code and dataset will be made available at:  \url{https://github.com/lvkailin0118/TouchThinker}.
\end{abstract}

\section{Introduction}
\label{Introduction}


Among human senses, touch is fundamental for perceiving and interacting with the physical world. It provides physical cues that vision alone cannot reliably capture, including material properties, surface texture, and contact state, thereby supporting physical reasoning and action decisions~\citep{tactile3,lyu2026tacreasoner,xiao2026tacexpert}. For example, when a person touches a hard grain of rice, they can infer that it is under-ripe; similarly, when touching a soft sponge, they can deduce that it is suitable for wiping based on the context. Therefore, incorporating tactile information into commonsense reasoning frameworks and developing tactile reasoning in open-world scenarios are crucial for advancing embodied intelligence.

Several recent studies have begun to integrate tactile perception with large language models~\citep{octopi,octopi15,vtv,stola}, enabling robots to perform tactile understanding tasks under natural language instructions. Despite these advances, existing approaches still struggle to achieve reliable and robust tactile reasoning in open-world environments, primarily due to limitations in data and representation. (1) \textbf{On the data side}, existing public tactile reasoning datasets remain limited in both format and scale~\citep{octopi,octopi15,vtv}, hindering models’ ability to perform tactile reasoning in open-world environments. Specifically, most rely on limited predefined attributes and question-answering templates, lacking causal reasoning supervision from tactile observations to physical properties and further to physical commonsense, which can induce hallucinations~\citep{huanjue}. Moreover, these datasets typically cover only one to three sensor types, making it difficult for models to distinguish sensor-specific representational biases from physical property variations shared across sensors, thereby limiting their ability to learn transferable tactile commonsense and cross-sensor physical reasoning.  (2) \textbf{On the representation side}, existing methods still struggle to effectively model tactile features. First, compared with visual signals, tactile interaction streams typically contain numerous static and transitional frames, whereas task-relevant attributes are localized within only a few highly informative segments~\citep{visuotactilesurvey}, resulting in substantial redundancy. In addition, tactile signals are inherently action-specific: pressing primarily reveals hardness, sliding captures friction, and rotation exposes texture. However, existing methods typically use uniform sampling or full-frame encoding, treating all frames equally. This makes it difficult to suppress noise from low-information frames and explicitly align action types with question semantics, resulting in inefficient representations with limited semantic expressiveness.

To address these challenges, we introduce TouchThinker, an open-world tactile-language framework that advances both data and representation. On the data side, we construct TouchThinker-1M, which integrates multi-source visuo-tactile data and applies systematic preprocessing, annotation unification, and question-answer expansion to provide semantically consistent supervision for tactile reasoning. On the representation side, we design an action-aware modeling mechanism that first fuses tactile tokens under question guidance and then performs action-aware Gaussian MoE modeling to identify query-relevant action segments and reduce representational redundancy. Based on it, we further develop a two-stage tactile-language training paradigm that aligns tactile evidence with language-based reasoning in the LLM. Experiments across multiple benchmarks, including the VTV-150K~\citep{vtv} and our newly constructed TouchThinker-Bench, demonstrate that TouchThinker achieves higher reasoning accuracy than state-of-the-art methods and supports reliable open-world tactile inference.

In summary, our contributions are as follows:
\begin{itemize}[leftmargin=*, itemsep=0pt, parsep=0pt, topsep=2pt, partopsep=0pt]
    \item \textbf{Framework}: We propose \textbf{TouchThinker}, a tactile-language reasoning framework that aligns tactile cues with task semantics through an action-aware modeling mechanism, enabling efficient tactile representations.
    \item \textbf{Datasets}: We construct \textbf{TouchThinker-1M}, a million-scale, multi-source dataset that expands tactile data across multiple dimensions to support open-world tactile reasoning. We further introduce \textbf{TouchThinker-Bench}, a systematic benchmark covering diverse tactile sensors and task types for rigorous multi-dimensional evaluation of open-world tactile commonsense reasoning.
    \item \textbf{Practice}: TouchThinker outperforms existing tactile-language models across multiple mainstream datasets and subtasks, while demonstrating more reliable physical reasoning. These results highlight its potential for robotic interaction in open-world environments and provide new insights for future research on tactile intelligence.
\end{itemize}

\vspace{-1.2mm}
\section{Related work}
\label{Related work}
\vspace{-1mm}

\noindent \textbf{Tactile Sensing and Perception.} In recent years, tactile sensing has evolved from early capacitive arrays, piezoresistive sensors, and magnetic sensors to advanced vision based systems capable of capturing high resolution contact information~\citep{sensorreview,sensorreview2}. visuo-tactile sensors have attracted growing attention because they can record fine grained spatiotemporal deformation on contact surfaces. Representative systems include DIGIT, GelSight, and Tac3D~\citep{tactile3,lyu2026vq}. Building on these advances, many studies have used visuo-tactile sensing to infer multidimensional tactile properties, thereby supporting dexterous manipulation tasks such as material classification, grasping, and insertion~\citep{touchformer,omnivta,zhang2026restacvla}. Recent research has shifted toward representation learning for tactile data. UniTouch~\citep{unitouch} and T3~\citep{T3} adopt visual self supervised objectives for fine grained feature extraction, while UniT~\citep{unit} introduces VQGAN~\citep{vqgan} for compact tactile latent space modeling. To address sensor heterogeneity, several studies employ joint training and alignment strategies to encourage consistent representations across sensors~\citep{vtv,anytouch}. In contrast, we identify representational redundancy in tactile signals and emphasize the action specific nature of tactile information. Based on this observation, we aim to learn more efficient tactile representations and unlock their potential for complex tactile reasoning.

\noindent \textbf{Tactile Commonsense Reasoning.} Multimodal large language models jointly model language and visual information, substantially improving cross modal reasoning and reshaping research paradigms~\citep{mllmsurvey,llmsurvey,qwen}. While early work primarily focused on vision language models~\citep{vlm1,wu2026dual,lyu2026himemvln,luo2026survey,lyu2026instruction}, recent studies have begun to leverage the reasoning and understanding capabilities of large language models to model tactile signals, gradually establishing a touch language modeling paradigm for embodied interaction. Representative studies such as Octopi~\citep{octopi}, Octopi 1.5~\citep{octopi15}, and VTV-LLM~\citep{vtv} have introduced important methods and benchmark datasets for tactile reasoning. However, taken together, existing studies remain constrained by limited datasets (e.g., sensor, scale, and formats), and directly transferring vision-based methods overlooks the unique properties of tactile perception. Motivated by this gap, we curate and release TouchThinker-1M, the largest tactile commonsense reasoning dataset to date to the best of our knowledge. It substantially surpasses existing benchmarks by scaling the data across multiple dimensions, thereby providing a stronger data foundation. Coupled with our methodological improvements, TouchThinker enhances the practical applicability and generalization of tactile commonsense reasoning in open-world scenarios.

\begin{figure*}[!t] 
    \centering 
    \includegraphics[width=\textwidth]{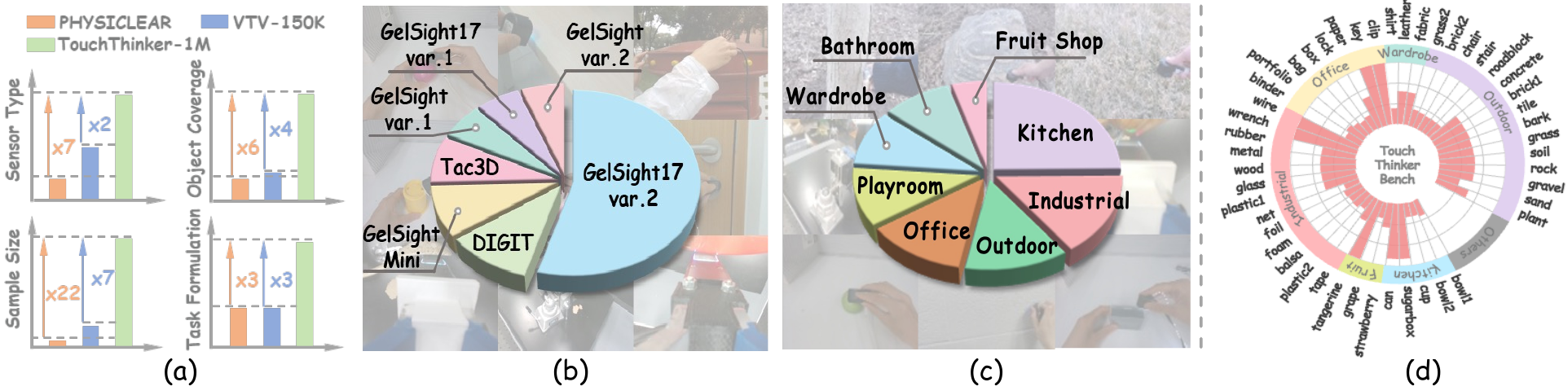}
    \vspace{-7mm}
    \caption{\textbf{Data statistics of TouchThinker-1M and TouchThinker-Bench.} (a) Scale comparison between TouchThinker-1M and representative tactile reasoning datasets in terms of sensor type, object coverage, sample size, and task formulation. (b) Distribution of tactile sensor types in TouchThinker-1M. (c) Distribution of scene categories in TouchThinker-1M. (d) Semantic taxonomy and category distribution of TouchThinker-Bench.} 
    \vspace{-3mm}
    \label{dataset_main} 
\end{figure*}

\vspace{-1mm}
\section{Curating TouchThinker-1M Dataset}
\label{Curating TouchThinker-1M Dataset}
\vspace{-1mm}

\textbf{Multi-Source Tactile Data Collection.} To address the limitations of existing tactile reasoning datasets in format and scale, which constrain models’ reasoning capabilities, we construct \textbf{TouchThinker-1M}, a large-scale, multi-source visuo-tactile dataset for open-world tactile commonsense reasoning. It systematically integrates mainstream visuo-tactile data sources from nine datasets, including representative sources such as TacQuad~\citep{anytouch}, PHYSICLEAR~\citep{octopi}, and VTV-150K~\citep{vtv}, comprising approximately \textbf{1M frames} in total. The incorporation of heterogeneous sources substantially broadens the distribution of tactile signals and provides a foundation for learning generalizable tactile semantic representations. As shown in Figure~\ref{dataset_main}, TouchThinker-1M significantly improves the scale and diversity of tactile data along five complementary dimensions: sensor type, interaction action, object coverage, sample size, and task formulation. Dataset details are provided in the \textbf{Appendix}~\ref{Source Data of TouchThinker-1M}.

\noindent \textbf{Tactile Attribute Annotation and Unification.} Since annotation schemas vary across tactile datasets, directly merging them would lead to an inconsistent semantic space and hinder learning of stable tactile concepts. To address this, we therefore adapt the VTV-150K~\citep{vtv} schema into a unified four-dimensional tactile attribute space covering Hardness, Protrusion, Elasticity, and Friction. For data with existing annotations, we map the original labels to this unified schema; For unannotated data, five annotators independently assign three-level labels for Hardness, Protrusion, Elasticity, and Friction by jointly considering object appearance, tactile observations, and contact-induced deformation. The annotations are cross-validated, and disagreements are resolved through adjudication. Samples with insufficient tactile evidence are excluded. The inter-annotator agreement scores~\citep{octopi} for Hardness, Protrusion, Elasticity, and Friction are 0.907, 0.924, 0.929, and 0.875, respectively, indicating high annotation consistency.

\noindent \textbf{Standardized Processing of Tactile Videos.} We standardize all samples into a unified video format to capture tactile interaction dynamics during contact. For existing tactile videos, we retain valid contact intervals and remove noncontact, noisy, or redundant segments. For static tactile images, we construct short video sequences by organizing frames according to object category and contact state to capture deformation evolution. All videos are further processed through contact region cropping, frame rate resampling, interpolation, temporal truncation, and length normalization, yielding clips of approximately 6 to 8 seconds that allow TouchThinker-1M to encode diverse tactile dynamics in a consistent format.

\noindent \textbf{Question Answer Format Expansion.} To reduce shallow semantic learning from templated tactile question answering~\citep{octopi,vtv}, TouchThinker-1M additionally introduces two complementary formats: chain-of-thought reasoning and open-ended question answering. Tactile chain-of-thought instances require an intermediate rationale and a final answer using {\small \texttt{<think>...</think><answer>...</answer>}}, where {\small \texttt{<think></think>}} records tactile evidence and reasoning paths, and {\small \texttt{<answer></answer>}} provides attribute judgments or commonsense conclusions. Open-ended tactile question answering covers free form description, comparative reasoning, attribute explanation, interaction prediction, and open-world decision making, enabling more realistic tactile language reasoning. Construction details are provided in the \textbf{Appendix}~\ref{Tactilc Instruction Data Synthesis}.



\section{TouchThinker-Bench}
Existing tactile reasoning benchmarks are often constrained by limited object categories, templated question answering formats, and restricted sensor coverage~\citep{octopi,vtv}, making it difficult to assess whether models acquire transferable tactile semantics. To comprehensively evaluate tactile commonsense reasoning in open-world scenarios, we introduce TouchThinker-Bench. It is designed for more realistic open-world evaluation and incorporates unseen objects, unseen sensors, and diverse tasks, thereby enabling systematic assessment of tactile understanding, commonsense reasoning, and generalization across sensors.

\noindent \textbf{Dataset creation and selection.} We first select samples from TouchThinker-1M using an object-level 6:1 train-test split. To evaluate cross-sensor generalization, we further incorporate self-collected data and several additional datasets~\citep{anytouch,T3}, covering three sensor types unseen during training. These sensors differ substantially in imaging mechanisms and exhibit intrinsically distinct data characteristics. Then, we apply data processing procedures similar to those described in Section~\ref{Curating TouchThinker-1M Dataset} and conduct manual verification, yielding a total of 82 test objects across the unseen-object and unseen-sensor settings, with their distribution shown in Figure~\ref{dataset_main}(d). Further dataset collection processes and dataset details are provided in the \textbf{Appendix}~\ref{Source Data of TouchThinker-Bench}.

\noindent \textbf{Task Taxonomy.} TouchThinker-Bench comprises three core task categories. \textit{(1) Basic tactile property understanding.} This task requires models to identify fundamental tactile attributes, including hardness, roughness, elasticity, and friction. \textit{(2) Basic tactile reasoning.} Following prior benchmarks such as Octopi~\citep{octopi} and VTV~\citep{vtv}, this task covers surface feature distinction (SFD), surface optimality identification (SOI), object sensation correlation (OSC), and tactile scenario analysis (TSA), enabling stable quantitative comparison. \textit{(3) Open-ended tactile commonsense reasoning.} It requires models to generate free-form responses for tactile phenomenon description, attribute explanation, and related reasoning tasks, assessing whether they can integrate tactile evidence with real-world commonsense rather than merely matching fixed templates. Detailed task definitions and examples are provided in the \textbf{Appendix}~\ref{Task Types}.

\begin{figure*}[!t] 
    \centering 
    \includegraphics[width=\textwidth]{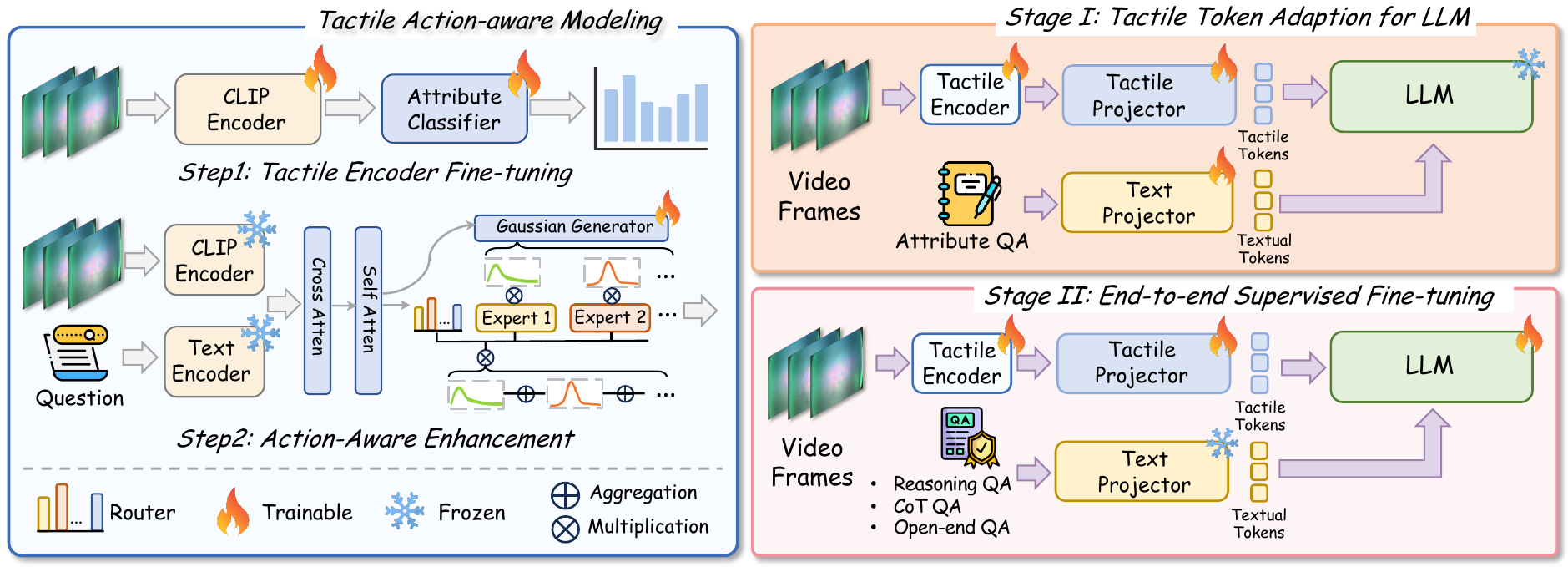}
    \vspace{-4mm}
    \caption{\textbf{Overview of the proposed TouchThinker framework and its training pipeline.}}
    \label{pipline}
\end{figure*}

\section{Methodology}
\label{Methodology}

\subsection{Tactile Action-aware Modeling}
\label{Action-aware Tactile Encoder}

As discussed in Section~\ref{Introduction}, tactile signals contain substantial redundancy, with task-relevant evidence concentrated in only a few action segments. Modeling all frames indiscriminately amplifies representational redundancy and hinders downstream reasoning. To address this issue, we propose a tactile action-aware modeling mechanism that localizes query-relevant subintervals and extracts meaningful tactile representations, as shown in Figure~\ref{pipline}.

\noindent \textbf{Question-Guided Token Fusion.} Given a tactile video $V=\{I_t\}_{t=1}^{T}$, we first fine-tune a base tactile backbone $f_{\mathrm{tac}}$ initialized from ViFi-CLIP~\citep{clip}. Specifically, we attach an attribute classifier and optimize a cross-entropy loss to predict tactile attributes, including hardness, protrusion, elasticity, and friction. After training, $f_{\mathrm{tac}}$ is frozen and used to extract frame-wise tactile features, yielding a tactile token sequence $F=[f_1,\ldots,f_T]\in\mathbb{R}^{T\times d}$. To incorporate task-specific priors, we encode the question $q$ with a frozen text encoder $f_{\mathrm{txt}}(\cdot)$, obtaining word-level features $Q_w\in\mathbb{R}^{L\times d_t}$ and a sentence-level feature $q_s\in\mathbb{R}^{d_t}$, which are projected into the tactile space as $\tilde{Q}_w=Q_w W_w$ and $\tilde{q}_s=q_s W_s$. We then use tactile tokens as queries and word-level question features as keys and values for cross-modal attention, followed by self-attention:
\begin{equation}
F_{\mathrm{qa}}
=
\mathrm{SelfAttn}
\left(
\mathrm{CrossAttn}
\left(
F,\tilde{Q}_w,\tilde{Q}_w
\right)
\right),
\end{equation}
The output $F_{\mathrm{qa}}\in\mathbb{R}^{T\times d}$ provides question-aware tactile representations that align temporal tactile cues with task semantics while suppressing irrelevant noise (e.g., non-contact segments), thereby reducing tactile representational redundancy.

\noindent \textbf{Action-aware Gaussian Temporal MoE.} To further focus on reasoning-relevant tactile evidence, we adapt the MoE framework by introducing a set of tactile experts, each specialized in distinct action patterns.
Rather than discrete experts~\citep{deepseekmoe,moe2}, we use context-dependent routing weights to softly combine their contributions, capturing the continuous dynamics of tactile interactions. Specifically, given the question-aware tactile tokens $F_{\mathrm{qa}}=[f_{\mathrm{qa},1},\ldots,f_{\mathrm{qa},T}]$, a question-driven router predicts the expert weights $\pi=\{\pi_k\}_{k=1}^{K}$ from the sentence-level question representation $\tilde{q}_s$. Each expert aggregates local tactile evidence through a Gaussian temporal window:
\begin{equation}
f_{\mathrm{moe}}
=
\sum_{k=1}^{K}\pi_k
\sum_{t=1}^{T}
\alpha_{k,t} E_k(f_{\mathrm{qa},t}),
\label{eq:temporal_moe}
\end{equation}
\begin{equation}
\alpha_{k,t}
=
\mathrm{Softmax}_{t}
\left(
-\frac{(\hat{\tau}_t-\mu_k)^2}{2\sigma_k^2}
\right),
\label{eq:temporal_weight}
\end{equation}
Here, $\alpha_{k,t}$ denotes the temporal attention weight assigned by the $k$-th expert to the $t$-th token, and $\hat{\tau}_t \in [0,1]$ denotes the normalized temporal position. The parameters $\mu_k$ and $\sigma_k$ denote the center and width of the temporal window for the $k$-th expert, respectively. 
They are generated from the $D$-dimensional question representation through a linear layer, where the predicted center offset is added to an initial temporal center. This design enables the model to focus on relevant action segments and capture the action-specific nature of tactile signals, thereby supporting downstream tactile reasoning.

\subsection{Tactile-Language training paradigm}
\label{Tactile-Language training paradigm}

In the previous section, we introduced a tactile action-aware mechanism for generating more efficient tactile representations. We next propose a two-stage training paradigm that enables the LLM to integrate tactile and linguistic embeddings for multimodal reasoning and response generation. The training pipeline is illustrated in Figure~\ref{pipline}.

\noindent \textbf{Stage I: Tactile Token Adaption for LLM.} In this stage, we align tactile encoder outputs with the LLM text embedding space through a tactile-language adapter, enabling the LLM to interpret tactile inputs. Trained on TouchThinker-1M attribute question answering data, the adapter projects tactile representations into textual token embeddings while the LLM~\citep{qwen} remains frozen, focusing optimization on cross-modal alignment. Given a tactile video $V$ and a text prompt $p$, the tactile embedding is computed as
\begin{equation}
E_V = \mathrm{Proj}(\mathrm{Enc}(V, p)),
\end{equation}
where $\mathrm{Enc}$ denotes the tactile encoder and $\mathrm{Proj}$ denotes
the tactile-language adapter. The LLM then generates a response $\hat{Y} = \mathrm{LLM}_{\phi}(E_V, p)$ based on the tactile embedding and the text prompt. The model is trained by minimizing the autoregressive cross entropy loss:
\begin{equation}
\mathcal{L}_{\mathrm{ce}}
=
-\mathbb{E}_{(V,p,Y)\sim \mathcal{D}_{\mathrm{align}}}
\sum_{i=1}^{M}
\log \pi_{\theta}(y_i \mid E_V, p, y_{<i}),
\end{equation}
where $\mathcal{D}_{\mathrm{align}}$ is the attribute question answering dataset. $Y=[y_1,\ldots,y_M]$ is the target answer sequence, $y_i$ is the target token at position $i$, $y_{<i}$ denotes the preceding target tokens, and $\pi_{\theta}$ denotes the predictive distribution. We use teacher forcing during training to stabilize tactile-text alignment.

\noindent \textbf{Stage II: End-to-end Supervised Fine-Tuning.} In this stage, we perform LoRA-based supervised fine-tuning on diverse instruction data, including chain-of-thought and open-ended question answering, to elicit evidence-driven and logically consistent tactile reasoning. To preserve stable perceptual representations, we freeze the tactile encoder and update only the
tactile-language adapter and the LoRA parameters inserted into the LLM
self-attention layers. Each instruction instance is represented as $(V,p,Y)$, where $V$ is a tactile video, $p$ is the text prompt, and $Y$ is the target output. For chain-of-thought instances, $Y$ follows the structured format
{\small \texttt{<think>...</think><answer>...</answer>}}, where the reasoning
trace $T$ precedes the final answer $A$. For open-ended question answering,
$Y$ is an open-format natural-language response. Under multi-format supervision, the model aligns dynamic tactile evidence with language-based reasoning. The supervised fine-tuning objective is defined as
\begin{equation}
\mathcal{L}_{\mathrm{sft}}
=
-\mathbb{E}_{(V,p,Y)\sim \mathcal{D}_{\mathrm{inst}}}
\sum_{i=1}^{M}
\log \pi_{\theta}(y_i \mid E_V,p,y_{<i}),
\end{equation}
where $\mathcal{D}_{\mathrm{inst}}$ denotes the instruction dataset containing
attribute question answering, chain-of-thought, and open-ended question
answering instances. This stage strengthens alignment between tactile
perception and language reasoning, improving generalization in open-world
tactile question answering.

\section{Experiments}
\label{Experiments}

\subsection{Settings}
\label{Settings}

\noindent \textbf{Datasets.} We conduct experiments on two benchmarks: the recently recognized VTV-150K~\citep{vtv} benchmark and our self-constructed TouchThinker-Bench. All test sets contain objects unseen during training, enabling comprehensive evaluation across diverse tactile reasoning tasks. In addition, TouchThinker-Bench further includes unseen sensors to assess cross-sensor generalization.

\noindent \textbf{Implementation.} All models are trained on four NVIDIA A100-80G GPUs with a batch size of 16. Tactile-language training follows a two-stage paradigm: Stage I trains the Tactile-Language Adapter on TouchThinker-1M attribute question-answering data using AdamW with a learning rate of $2 \times 10^{-4}$; Stage II fine-tunes the adapter and the LLM LoRA parameters on more complex question-answering data, including chain-of-thought and open-ended question answering, with the same optimizer and learning rate. 
We set the LoRA scaling factor to 256, rank to 128, and maximum training steps to 10,000. 
The LLM backbone is Qwen2.5, including 7B and 14B~\citep{qwen}. Additional details are provided in the \textbf{Appendix}~\ref{Configuration and Hyperparameters}.

\begin{table*}[!t]
\centering
\caption{\textbf{Performance comparison of TouchThinker with state-of-the-art methods on the VTV-150K dataset.} The evaluation covers multiple tasks, with results reported in percentages (\%); boldface denotes the best performance. To ensure fair comparison and validate the effectiveness of our method, TouchThinker is trained and evaluated on VTV-150K following the same protocol~\citep{vtv}, and results are reported as the average over multiple runs.}
\vspace{-2mm}
\resizebox{\textwidth}{!}{
\begin{tabular}{l|ccccc|cccc|c}
\hline
Models & Hardness & Protrusion & Elasticity & Friction & Combined & SFD & SOI & OSC & TSA & Average \\
\hline
GPT-4o & 34.7 & 32.6 & 32.6 & 18.7 & 2.1 & 40.9 & 38.4 & 16.6 & 36.0 & 28.0 \\
Gemini-2.5-Pro-Exp & 36.2 & 34.7 & 39.1 & 21.0 & 4.3 & 42.6 & 29.4 & 18.5 & 40.0 & 29.5 \\
LLaVA-OneVision-7B & 27.5 & 32.6 & 26.0 & 20.2 & 0.7 & 40.9 & 28.2 & 11.7 & 30.0 & 24.2 \\
LLaVA-Video-Qwen2-7B & 30.4 & 29.7 & 28.9 & 18.1 & 2.1 & 33.6 & 29.4 & 17.2 & 36.0 & 25.0 \\
InternVL2.5-VL-8B & 18.1 & 23.9 & 21.0 & 13.7 & 0.0 & 24.5 & 17.9 & 11.1 & 24.0 & 17.1 \\
VideoLLaMA3-7B & 15.2 & 21.7 & 14.4 & 10.8 & 0.0 & 11.4 & 12.8 & 7.4 & 20.0 & 12.6 \\
Qwen2.5-VL-7B & 25.3 & 28.9 & 17.3 & 15.9 & 1.4 & 22.9 & 28.2 & 16.0 & 30.0 & 20.6 \\
\hline
VTV-LLM-7B & 73.9 & 75.0 & 67.3 & 56.5 & 35.6 & 71.3 & 57.6 & 43.2 & 64.0 & 60.4 \\
VTV-LLM-14B & 72.1 & 78.2 & 68.1 & 52.8 & 38.2 & 72.1 & 59.7 & 45.9 & 72.0 & 62.1 \\
\hline
\rowcolor{lightblue} \textbf{TouchThinker-7B} 
& 79.1 
& \textbf{80.8} 
& 75.3 
& \textbf{63.4} 
& 40.7 
& 78.9 
& 64.2 
& 50.7 
& 74.0 
& 67.5 \\

$\Delta$ (vs. VTV-LLM-7B) 
& \color{mygreen}+5.2 
& \color{mygreen}+5.8 
& \color{mygreen}+8.0 
& \color{mygreen}+6.9 
& \color{mygreen}+5.1 
& \color{mygreen}+7.6 
& \color{mygreen}+6.6 
& \color{mygreen}+7.5 
& \color{mygreen}+10.0 
& \color{mygreen}+7.0 \\

\textbf{TouchThinker-14B} 
& \textbf{81.2} 
& 77.1 
& \textbf{78.2} 
& 61.5 
& \textbf{46.4} 
& \textbf{80.5} 
& \textbf{68.3} 
& \textbf{52.9} 
& \textbf{80.0} 
& \textbf{69.6} \\
\hline
\end{tabular}
}
\label{tab1}
\label{vtv_task}
\end{table*}

\noindent \textbf{Evaluation Metrics.} For basic tactile property understanding and reasoning, we report subtask-level accuracy by exactly matching the final conclusions, without considering reasoning processes or semantic equivalence. For open-ended questions, we measure semantic similarity with METEOR~\citep{meteor} and use GPT-5~\citep{gpt5} and DeepSeek-V4~\citep{deepseekv4} for multidimensional evaluation. 

\vspace{-2mm}
\subsection{Main Results}
\label{Main Results}


The main experimental results demonstrate that TouchThinker scales tactile commonsense reasoning to open-world scenarios with three key \textit{\textbf{adv}}antages: \textit{\textbf{[Adv.1]}} improved tactile property prediction and basic tactile commonsense reasoning, \textit{\textbf{[Adv.2]}} accurate and coherent open-ended response generation, and \textit{\textbf{[Adv.3]}} robust generalization to unseen sensors and objects.


\noindent \textit{\textbf{[Adv.1]}} To ensure fair comparison and validate the effectiveness of our method, we follow the VTV-LLM evaluation protocol~\citep{vtv} and quantitatively compare TouchThinker with state-of-the-art models on 500 question-answer pairs sampled from VTV-150K. The results are summarized in Table~\ref{vtv_task}. On Tactile Feature Analysis task, TouchThinker-7B outperforms VTV-LLM-7B by 5.2\%, 5.8\%, 8.0\%, and 6.9\% in hardness, protrusion, elasticity, and friction prediction, respectively, demonstrating the effectiveness of its action-aware tactile representations. We further evaluate TouchThinker on basic tactile commonsense reasoning tasks, including SFD, SOI, OSC, and TSA. Compared with VTV-LLM-7B, TouchThinker consistently achieves performance gains. Notably, TouchThinker-7B also surpasses VTV-LLM-14B in overall performance and on most subtasks despite using substantially fewer parameters, highlighting its efficient tactile representation learning and stronger reasoning capability.

\begin{figure}[!t]
    \centering
    \includegraphics[width=\linewidth]{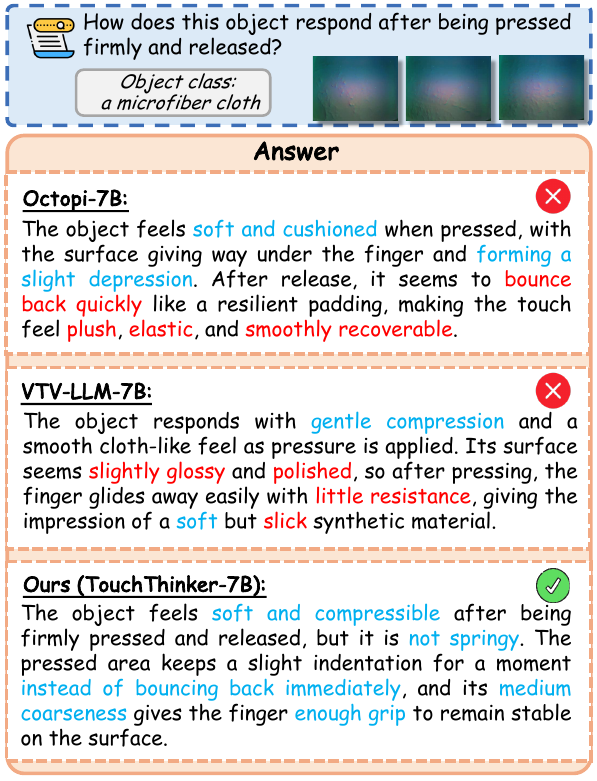}
    \caption{Qualitative comparison of responses from TouchThinker and other models. Errors are highlighted in \textbf{\textcolor{red}{red}}, while accurate content is marked in \textbf{\textcolor{blue}{blue}}.}
    \label{reasoning_example}
\end{figure}

\noindent \textit{\textbf{[Adv.2]}} Prior methods perform reasonably on tactile property prediction and templated reasoning, but their reliance on fixed answer spaces and exact matching limits the evaluation of open-world tactile explanation and reasoning. Following the implementation details described in Section~\ref{Settings}, we further evaluate TouchThinker on TouchThinker-Bench for open-ended tactile question answering, using METEOR~\citep{meteor} and LLM-based evaluation. As shown in Table~\ref{tactilebench_subtasks}, Large-scale multi-source data and diversified QA training enable TouchThinker to reduce shallow semantic matching and spurious associations, producing responses that are better grounded in tactile evidence and more logically consistent. The qualitative comparison in Figure~\ref{reasoning_example} further shows that TouchThinker better interprets tactile signals, connects physical properties with real-world commonsense, and performs open-world tactile reasoning.

\begin{table*}[!t]
\centering
\caption{\textbf{Subtask performance comparison on TouchThinker-Bench.} We evaluate the models’ open-ended tactile question-answering capabilities. The best results are in \textbf{bold}, and the second-best ones are \underline{underlined}. TAU: Touch Attribute Understanding. TIU: Touch Interaction Understanding. TKU: Touch Knowledge Reasoning.}
\resizebox{\textwidth}{!}{
\begin{tabular}{l|ccc|ccc|ccc}
\hline
\multirow{2}{*}{Model} 
& \multicolumn{3}{c|}{TAU} 
& \multicolumn{3}{c|}{TIU} 
& \multicolumn{3}{c}{TKU} \\
\cline{2-10}
& METEOR & GPT-5 & DeepSeek-V4 
& METEOR & GPT-5 & DeepSeek-V4 
& METEOR & GPT-5 & DeepSeek-V4 \\
\hline
Octopi-7B~\citep{octopi}
& 26.47 & 6.45 & 6.42 
& 25.26 & 6.89 & 6.94 
& 20.84 & 6.23 & 6.14 \\

Octopi-13B~\citep{octopi} 
& \underline{31.43} & \underline{7.58} & \underline{7.87} 
& 24.41 & \underline{7.11} & \textbf{7.79}
& \underline{26.76} & \underline{6.39} & \underline{6.87} \\

VTV-LLM-7B~\citep{vtv}
& 27.93 & 6.53 & 6.64 
& \underline{27.45} & 6.87 & 6.91 
& 22.17 & 6.19 & 6.12 \\

TouchThinker-7B (Ours) 
& \textbf{34.06} & \textbf{8.17} & \textbf{8.33} 
& \textbf{28.71} & \textbf{7.49} & \underline{7.21} 
& \textbf{27.43} & \textbf{7.87} & \textbf{7.81} \\
\hline
\end{tabular}
}
\label{tactilebench_subtasks}
\end{table*}

\begin{table*}[!t]
\centering
\caption{\textbf{Performance comparison on TouchThinker-Bench under unseen sensor and object.} Other models suffer substantial performance degradation due to limited generalization, whereas our method maintains robustness.}
\vspace{-2mm}
\resizebox{\textwidth}{!}{
\begin{tabular}{l|ccccc|cccc|c}
\hline
Models & Hardness & Protrusion & Elasticity & Friction & Combined & SFD & SOI & OSC & TSA & Average \\
\hline
Octopi-7B  & 43.2 & 47.1 & 39.5 & 32.6 & 13.9 & 31.4 & 33.9 & 22.4 & 48.0 & 34.7 \\
Octopi-13B & 44.8 & 46.5 & 41.9 & 36.1 & 22.1 & 35.5 & 38.7 & 30.6 & 56.0 & 39.1 \\
VTV-LLM-7B & 59.2 & 60.1 & 62.8 & 46.5 & 28.4 & 51.6 & 43.5 & 39.7 & 52.0 & 49.3 \\
\hline
\rowcolor{lightblue} \textbf{TouchThinker-7B} & \textbf{68.3} & \textbf{69.7} & \textbf{70.6} & \textbf{51.7} & \textbf{37.2} & \textbf{68.2} & \textbf{52.1} & \textbf{47.2} & \textbf{62.0} & \textbf{58.6} \\
\hline
\end{tabular}
}
\vspace{-2mm}
\label{tab3}
\end{table*}

\begin{table}[!t]
\centering
\renewcommand{\arraystretch}{1.1}
\caption{Ablation study of TouchThinker components.}
\vspace{-2mm}
\label{tab4}
\resizebox{\columnwidth}{!}{
\begin{tabular}{l|cccc|c}
\hline
Settings & SFD & SOI & OSC & TSA & Average \\
\hline
w/o Action-aware Modeling & 72.9 & 57.1 & 46.3 & 70.0 & 61.6 \\
w/o stage 1 & 65.3 & 58.7 & 41.2 & 68.0 & 58.3 \\
w/o stage 2 & 61.1 & 51.3 & 34.9 & 66.0 & 53.3 \\
\textbf{Ours} & \textbf{78.9} & \textbf{64.2} & \textbf{50.7} & \textbf{74.0} & \textbf{67.0} \\
\hline
\end{tabular}
}
\vspace{-2mm}
\label{tableab}
\end{table}

\noindent \textit{\textbf{[Adv.3]}} We evaluate state-of-the-art Touch-Language models on TouchThinker-Bench for tactile property prediction and basic tactile reasoning. Table~\ref{tab3} shows that TouchThinker remains robust to unseen sensors and objects, while Octopi and VTV-LLM degrade substantially. This indicates stronger adaptation to heterogeneous sensor imaging mechanisms, signal distributions, and object properties. This advantage stems from the large-scale, multi-source construction of TouchThinker-1M, which enables models to learn sensor-invariant tactile semantics rather than sensor-specific appearance biases, thereby improving generalization to unseen sensing platforms and open-world objects.

\begin{figure}[!t] 
    \centering 
    \includegraphics[width=\columnwidth]{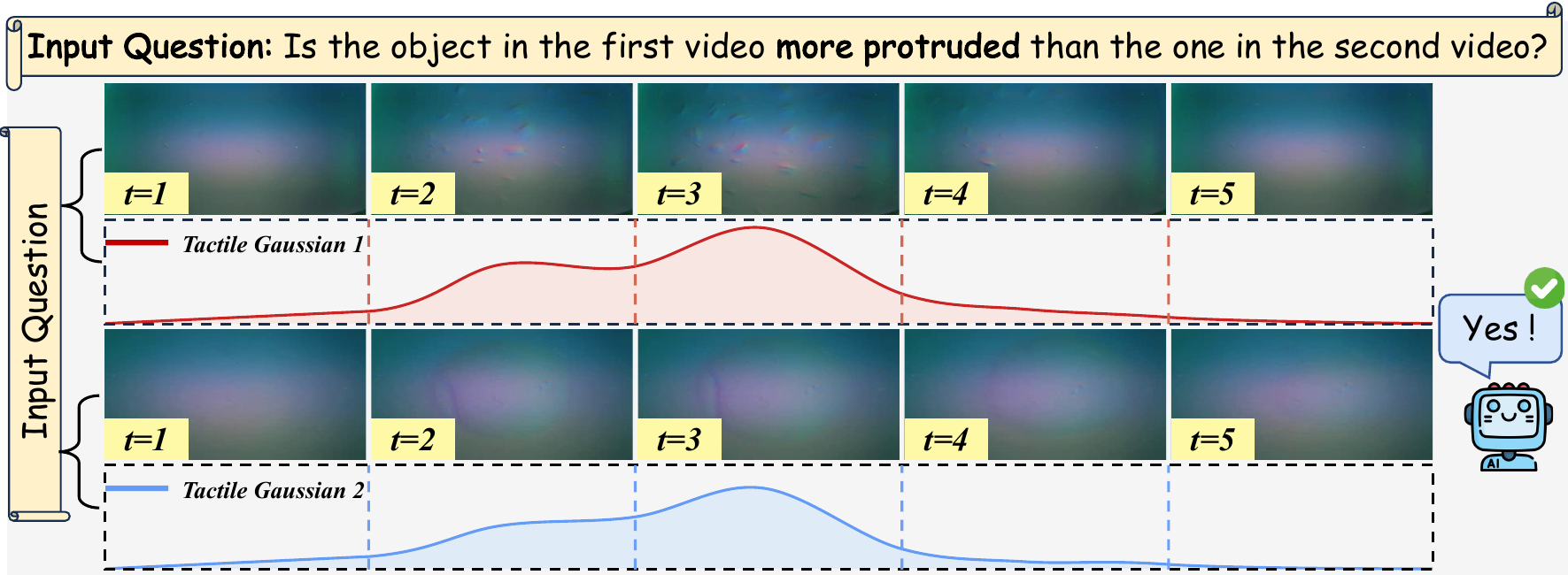}
    \vspace{-5mm}
    \caption{Visualization of action-aware temporal weights from tactile Gaussian experts, integrated to focus on question-relevant frames for accurate reasoning.}
    \vspace{-4mm}
    \label{TouchThinker_Gaussian}
\end{figure}

\noindent \textbf{Impact of the Action-aware Modeling.} As shown in Table~\ref{tableab}, removing the action-aware mechanism leads to consistent performance degradation on both attribute prediction and reasoning tasks, due to the inherent redundancy and action specificity of tactile signals. In contrast, our action-aware mechanism optimizes tactile representations. To intuitively illustrate this effect, we present visualizations in Figure~\ref{TouchThinker_Gaussian}. When the question is related to the protrusion attribute, the action-aware mechanism dynamically localizes pressing-related action segments, yielding efficient tactile representations and enhancing tactile understanding.

\noindent \textbf{Impact of the Training Paradigm.} Table~\ref{tableab} validates our two-stage training paradigm through ablation studies on VTV-150K. Removing Stage I leads to misalignment between tactile and textual tokens, reducing the average performance to 58.3\%. Omitting Stage II removes end-to-end fine-tuning, destabilizes LLM token generation, and causes a sharp performance drop to 53.3\%. These results confirm that progressive training is crucial for improving model capability. 

\section{Conclusion}
\label{Conclusion}

In this work, we present TouchThinker, an open-world tactile-language reasoning framework that advances tactile reasoning from both data and methodological perspectives. First, we construct TouchThinker-1M, a million-scale multi-source visuo-tactile dataset, and TouchThinker-Bench, a realistic benchmark for evaluating tactile understanding, commonsense reasoning, and cross-sensor generalization. Then, TouchThinker employs a tactile action-aware modeling mechanism that combines question-guided fusion with action-aware Gaussian Temporal MoE to align tactile cues with task semantics, reduce redundancy, and focus on reasoning-relevant action segments. A two-stage tactile-language training paradigm further aligns tactile evidence with LLM-based reasoning. Extensive experiments show that TouchThinker outperforms existing tactile-language models across benchmarks, with stronger robustness to unseen objects and sensors. These results highlight the value of scalable tactile data and action-aware dynamics for reliable open-world tactile reasoning.

\section{Limitations}
\label{Limitations}
Although TouchThinker demonstrates promising capability in tactile commonsense reasoning, several limitations remain. First, its supervision is constrained by the diversity of annotated tactile attributes. The current schema covers common and complementary attributes, including hardness, protrusion, elasticity, friction, and their combinations, while real-world tactile perception involves additional properties such as malleability and prickliness. This may limit comprehensive tactile understanding in open-world settings. Second, our current implementation focuses mainly on short-term tactile interactions. Since TouchThinker-1M primarily consists of 6–8-second contact clips, extending the framework to long-horizon tactile manipulation remains important for more complex tasks~\citep{tactime,tactilelong}. Third, TouchThinker is evaluated with 7B and 14B LLM backbones, whose computational cost increases with model scale and may hinder deployment on resource-constrained robotic platforms. Developing lightweight and accelerated variants is a promising direction for future work.

\section{Broader Impacts and Ethics Statement}
\label{Broader Impacts and Ethics Statement}
This work advances open-world tactile commonsense reasoning by aligning language models with tactile perception, enabling embodied agents to acquire more reliable and interpretable physical understanding. It aligns with broader NLP interests in multimodal grounding, human-AI interaction, trustworthy evaluation, and responsible data curation. Potential benefits include improved assistive technologies for visually impaired individuals and applications in manufacturing, quality control, and robotic manipulation~\citep{embodiedtac}. We also acknowledge potential ethical risks. (1) Tactile data may encode fine-grained information about object materials, interaction patterns, or usage contexts, raising privacy and sensitive-information concerns if improperly collected, stored, or released. (2) Although TouchThinker-1M integrates multi-source tactile data, its sensor types, object categories, and attribute space cannot fully capture real-world variability, which may lead to generalization biases for low-resource sensors, rare materials, or safety-critical scenarios~\citep{multitrust}. To mitigate these risks, we emphasize lawful data sources, standardized annotation, cross-sensor evaluation, and open-world generalization analysis, and recommend further real-world validation of safety and feasibility before deployment.

\bibliography{custom}

@article{tactile3,
  title={A comprehensive review of robot intelligent grasping based on tactile perception},
  author={Li, Tong and Yan, Yuhang and Yu, Chengshun and An, Jing and Wang, Yifan and Chen, Gang},
  journal={Robotics and Computer-Integrated Manufacturing},
  volume={90},
  pages={102792},
  year={2024},
  publisher={Elsevier}
}

@article{octopi,
  title={Octopi: Object property reasoning with large tactile-language models},
  author={Yu, Samson and Lin, Kelvin and Xiao, Anxing and Duan, Jiafei and Soh, Harold},
  journal={arXiv preprint arXiv:2405.02794},
  year={2024}
}

@article{octopi15,
  title={Demonstrating the octopi-1.5 visual-tactile-language model},
  author={Yu, Samson and Lin, Kelvin and Soh, Harold},
  journal={arXiv preprint arXiv:2507.09985},
  year={2025}
}

@inproceedings{clip,
  title={Learning transferable visual models from natural language supervision},
  author={Radford, Alec and Kim, Jong Wook and Hallacy, Chris and Ramesh, Aditya and Goh, Gabriel and Agarwal, Sandhini and Sastry, Girish and Askell, Amanda and Mishkin, Pamela and Clark, Jack and others},
  booktitle={International conference on machine learning},
  pages={8748--8763},
  year={2021},
  organization={PmLR}
}

@article{vtv,
  title={Universal visuo-tactile video understanding for embodied interaction},
  author={Xie, Yifan and Li, Mingyang and Li, Shoujie and Li, Xingting and Chen, Guangyu and Ma, Fei and Yu, Fei and Ding, Wenbo},
  journal={Advances in Neural Information Processing Systems},
  volume={38},
  pages={127864--127883},
  year={2026}
}

@article{lyu2026tacreasoner, title={TacReasoner: A Dynamic Tactile-Language Framework for Interactive Reasoning in Real-World Scenarios}, author={Lyu, Kailin and Wu, Di and Xiao, Long and Zeng, Jianning and He, Jianwei and Lin, Chang and Hu, Lianyu and Shu, Lin and Hao, Jie and Hao, Ce}, journal={arXiv preprint arXiv:2607.05131}, year={2026} }

@inproceedings{xiao2026tacexpert,
  title={TacExpert: A Pseudo-Temporal Mixture-of-Experts Framework for Open-Set Tactile Object Recognition},
  author={Xiao, Long and Lyu, Kailin and He, Jianwei and Zeng, Jianing and Shu, Lin and Hao, Jie},
  booktitle={2026 IEEE International Conference on Acoustics, Speech and Signal Processing (ICASSP)},
  pages={3761--3765},
  year={2026},
  organization={IEEE}
}

@article{lyu2026himemvln, title={Himemvln: Enhancing reliability of open-source zero-shot vision-and-language navigation with hierarchical memory system}, author={Lyu, Kailin and Wu, Kangyi and Li, Pengna and Hu, Xiuyu and Si, Qingyi and Miao, Cui and Yang, Ning and Wang, Zihang and Xiao, Long and Hu, Lianyu and others}, journal={arXiv preprint arXiv:2603.14807}, year={2026} }

@inproceedings{zhang2026restacvla,
  title={Feeling the Unexpected: ResTacVLA for Contact-Rich Manipulation via Residual Tactile Representation},
  author={Zhang, Pengwei and Xie, Bin and Meng, Xinpan and Guo, Xinyu and Hao, Ce and Deng, Fang and Cheng, Long and Wang, Tiancai},
  booktitle={IEEE/RSJ International Conference on Intelligent Robots and Systems (IROS)},
  year={2026}
}

@article{wu2026dual, title={Dual-Anchoring: Addressing State Drift in Vision-Language Navigation}, author={Wu, Kangyi and Li, Pengna and Lyu, Kailin and Lin, Xi and Zhao, Lin and He, Qingrong and Wang, Jinjun and Liu, Jianyi}, journal={arXiv preprint arXiv:2604.17473}, year={2026} }

@article{lyu2026vq, title={VQ-Touch: A Data-Efficient Tactile Generation Framework Across Sensors and Scenarios}, author={Lyu, Kailin and Xiao, Long and Zeng, Jianing and Wu, Di and Shu, Lin and Hao, Jie}, journal={arXiv preprint arXiv:2607.14728}, year={2026} }

@article{huanjue,
  title={From image to language: A critical analysis of visual question answering (vqa) approaches, challenges, and opportunities},
  author={Ishmam, Md Farhan and Shovon, Md Sakib Hossain and Mridha, Muhammad Firoz and Dey, Nilanjan},
  journal={Information Fusion},
  volume={106},
  pages={102270},
  year={2024},
  publisher={Elsevier}
}

@article{sensorreview,
  title={Tactile sensors: A review},
  author={Meribout, Mahmoud and Takele, Natnael Abule and Derege, Olyad and Rifiki, Nidal and El Khalil, Mohamed and Tiwari, Varun and Zhong, Jing},
  journal={Measurement},
  volume={238},
  pages={115332},
  year={2024},
  publisher={Elsevier}
}

@article{sensorreview2,
  title={Classification of vision-based tactile sensors: A review},
  author={Li, Haoran and Lin, Yijiong and Lu, Chenghua and Yang, Max and Psomopoulou, Efi and Lepora, Nathan F},
  journal={IEEE Sensors Journal},
  year={2025},
  publisher={IEEE}
}

@article{visuotactilesurvey,
  title={Visuotactile sensors with emphasis on gelsight sensor: A review},
  author={Abad, Alexander C and Ranasinghe, Anuradha},
  journal={IEEE Sensors Journal},
  volume={20},
  number={14},
  pages={7628--7638},
  year={2020},
  publisher={IEEE}
}

@article{omnivta,
  title={Omnivta: Visuo-tactile world modeling for contact-rich robotic manipulation},
  author={Zheng, Yuhang and Gu, Songen and Li, Weize and Zheng, Yupeng and Zang, Yujie and Tian, Shuai and Li, Xiang and Hao, Ce and Gao, Chen and Liu, Si and others},
  journal={arXiv preprint arXiv:2603.19201},
  year={2026}
}

@inproceedings{touchformer,
  title={TouchFormer: A Robust Transformer-based Framework for Multimodal Material Perception},
  author={Lyu, Kailin and Xiao, Long and Zeng, Jianing and Dong, Junhao and Liu, Xuexin and Zou, Zhuojun and Yang, Haoyue and Shu, Lin and Hao, Jie},
  booktitle={Proceedings of the AAAI Conference on Artificial Intelligence},
  volume={40},
  number={22},
  pages={18496--18504},
  year={2026}
}

@inproceedings{unitouch,
  title={Binding touch to everything: Learning unified multimodal tactile representations},
  author={Yang, Fengyu and Feng, Chao and Chen, Ziyang and Park, Hyoungseob and Wang, Daniel and Dou, Yiming and Zeng, Ziyao and Chen, Xien and Gangopadhyay, Rit and Owens, Andrew and others},
  booktitle={Proceedings of the IEEE/CVF Conference on Computer Vision and Pattern Recognition},
  pages={26340--26353},
  year={2024}
}

@article{T3,
  title={Transferable tactile transformers for representation learning across diverse sensors and tasks},
  author={Zhao, Jialiang and Ma, Yuxiang and Wang, Lirui and Adelson, Edward H},
  journal={arXiv preprint arXiv:2406.13640},
  year={2024}
}

@article{unit,
  title={UniT: Data efficient tactile representation with generalization to unseen objects},
  author={Xu, Zhengtong and Uppuluri, Raghava and Zhang, Xinwei and Fitch, Cael and Crandall, Philip Glen and Shou, Wan and Wang, Dongyi and She, Yu},
  journal={IEEE Robotics and Automation Letters},
  year={2025},
  publisher={IEEE}
}

@inproceedings{vqgan,
  title={Taming transformers for high-resolution image synthesis},
  author={Esser, Patrick and Rombach, Robin and Ommer, Bjorn},
  booktitle={Proceedings of the IEEE/CVF conference on computer vision and pattern recognition},
  pages={12873--12883},
  year={2021}
}

@article{anytouch,
  title={Anytouch: Learning unified static-dynamic representation across multiple visuo-tactile sensors},
  author={Feng, Ruoxuan and Hu, Jiangyu and Xia, Wenke and Gao, Tianci and Shen, Ao and Sun, Yuhao and Fang, Bin and Hu, Di},
  journal={arXiv preprint arXiv:2502.12191},
  year={2025}
}

@article{llmsurvey,
  title={Surveying the mllm landscape: A meta-review of current surveys},
  author={Li, Ming and Chen, Keyu and Bi, Ziqian and Liu, Ming and Song, Xinyuan and Jiang, Zekun and Wang, Tianyang and Peng, Benji and Niu, Qian and Liu, Junyu and others},
  journal={arXiv preprint arXiv:2409.18991},
  year={2024}
}

@article{mllmsurvey,
  title={A survey on multimodal large language models},
  author={Yin, Shukang and Fu, Chaoyou and Zhao, Sirui and Li, Ke and Sun, Xing and Xu, Tong and Chen, Enhong},
  journal={National Science Review},
  volume={11},
  number={12},
  pages={nwae403},
  year={2024},
  publisher={Oxford University Press}
}

@article{vlm1,
  title={A survey of vision-language pre-trained models},
  author={Du, Yifan and Liu, Zikang and Li, Junyi and Zhao, Wayne Xin},
  journal={arXiv preprint arXiv:2202.10936},
  year={2022}
}

@article{luo2026survey, title={A survey of large audio language models: Generalization, trustworthiness, and outlook}, author={Luo, Kaiwen and Zhou, Zhenhong and Wang, Leyan and Lin, Liang and Shao, Tianyu and Zhang, Yuanhe and Xiao, Yang and Li, Yuxuan and Yu, Miao and Lyu, Kailin and others}, journal={arXiv preprint arXiv:2605.20266}, year={2026} }

@techreport{lyu2026instruction, title={From Instruction Following to Cognitive Navigation: A Survey on the Evolution of Vision-and-Language Navigation}, author={Lyu, Kailin and Wu, Kangyi and Li, Pengna and Song, Wenxuan and Wu, Di and He, Jianwei and Chen, Junting and Yang, Ning and Han, Zebin and Luo, Kaiwen and others}, year={2026}, institution={CSPaper OpenPrint} }

@inproceedings{stola,
  title={STOLA: Self-Adaptive Touch-Language Framework for Tactile Commonsense Reasoning in Open-Ended Scenarios},
  author={Cheng, Ning and Xu, Jinan and Chen, Jialing and Fang, Bin and Han, Wenjuan},
  booktitle={Proceedings of the AAAI Conference on Artificial Intelligence},
  volume={40},
  number={22},
  pages={18198--18206},
  year={2026}
}

@article{tag,
  title={Touch and go: Learning from human-collected vision and touch},
  author={Yang, Fengyu and Ma, Chenyang and Zhang, Jiacheng and Zhu, Jing and Yuan, Wenzhen and Owens, Andrew},
  journal={arXiv preprint arXiv:2211.12498},
  year={2022}
}

@inproceedings{objectfolder,
  title={The objectfolder benchmark: Multisensory learning with neural and real objects},
  author={Gao, Ruohan and Dou, Yiming and Li, Hao and Agarwal, Tanmay and Bohg, Jeannette and Li, Yunzhu and Fei-Fei, Li and Wu, Jiajun},
  booktitle={Proceedings of the IEEE/CVF Conference on Computer Vision and Pattern Recognition},
  pages={17276--17286},
  year={2023}
}

@inproceedings{YCB-Slide,
  title={Midastouch: Monte-carlo inference over distributions across sliding touch},
  author={Suresh, Sudharshan and Si, Zilin and Anderson, Stuart and Kaess, Michael and Mukadam, Mustafa},
  booktitle={Conference on Robot Learning},
  pages={319--331},
  year={2023},
  organization={PMLR}
}

@article{Touch-Slide,
  title={Sparsh: Self-supervised touch representations for vision-based tactile sensing},
  author={Higuera, Carolina and Sharma, Akash and Bodduluri, Chaithanya Krishna and Fan, Taosha and Lancaster, Patrick and Kalakrishnan, Mrinal and Kaess, Michael and Boots, Byron and Lambeta, Mike and Wu, Tingfan and others},
  journal={arXiv preprint arXiv:2410.24090},
  year={2024}
}

@article{FeelSight,
  title={NeuralFeels with neural fields: Visuotactile perception for in-hand manipulation},
  author={Suresh, Sudharshan and Qi, Haozhi and Wu, Tingfan and Fan, Taosha and Pineda, Luis and Lambeta, Mike and Malik, Jitendra and Kalakrishnan, Mrinal and Calandra, Roberto and Kaess, Michael and others},
  journal={Science Robotics},
  volume={9},
  number={96},
  pages={eadl0628},
  year={2024},
  publisher={American Association for the Advancement of Science}
}

@inproceedings{HaTT,
  title={One hundred data-driven haptic texture models and open-source methods for rendering on 3D objects},
  author={Culbertson, Heather and Delgado, Juan Jos{\'e} L{\'o}pez and Kuchenbecker, Katherine J},
  booktitle={2014 IEEE haptics symposium (HAPTICS)},
  pages={319--325},
  year={2014},
  organization={IEEE}
}

@inproceedings{deepseekmoe,
  title={Deepseekmoe: Towards ultimate expert specialization in mixture-of-experts language models},
  author={Dai, Damai and Deng, Chengqi and Zhao, Chenggang and Xu, RX and Gao, Huazuo and Chen, Deli and Li, Jiashi and Zeng, Wangding and Yu, Xingkai and Wu, Yu and others},
  booktitle={Proceedings of the 62nd Annual Meeting of the Association for Computational Linguistics (Volume 1: Long Papers)},
  pages={1280--1297},
  year={2024}
}

@article{moe2,
  title={Moe-llava: Mixture of experts for large vision-language models},
  author={Lin, Bin and Tang, Zhenyu and Ye, Yang and Huang, Jinfa and Zhang, Junwu and Pang, Yatian and Jin, Peng and Ning, Munan and Luo, Jiebo and Yuan, Li},
  journal={IEEE Transactions on Multimedia},
  year={2026},
  publisher={IEEE}
}

@article{qwen,
  title={Qwen-image technical report},
  author={Wu, Chenfei and Li, Jiahao and Zhou, Jingren and Lin, Junyang and Gao, Kaiyuan and Yan, Kun and Yin, Sheng-ming and Bai, Shuai and Xu, Xiao and Chen, Yilei and others},
  journal={arXiv preprint arXiv:2508.02324},
  year={2025}
}

@inproceedings{meteor,
  title={METEOR: An automatic metric for MT evaluation with improved correlation with human judgments},
  author={Banerjee, Satanjeev and Lavie, Alon},
  booktitle={Proceedings of the acl workshop on intrinsic and extrinsic evaluation measures for machine translation and/or summarization},
  pages={65--72},
  year={2005}
}

@article{embodiedtac,
  title={Embodied tactile perception and learning},
  author={Liu, Huaping and Guo, Di and Sun, Fuchun and Yang, Wuqiang and Furber, Steve and Sun, Tengchen},
  journal={Brain Science Advances},
  volume={6},
  number={2},
  pages={132--158},
  year={2020},
  publisher={SAGE Publications Sage UK: London, England}
}

@article{multitrust,
  title={Multitrust: A comprehensive benchmark towards trustworthy multimodal large language models},
  author={Zhang, Yichi and Huang, Yao and Sun, Yitong and Liu, Chang and Zhao, Zhe and Fang, Zhengwei and Wang, Yifan and Chen, Huanran and Yang, Xiao and Wei, Xingxing and others},
  journal={Advances in Neural Information Processing Systems},
  volume={37},
  pages={49279--49383},
  year={2024}
}

@article{tactilelong,
  title={Tactile sensing in intelligent robotic manipulation--a review},
  author={Tegin, Johan and Wikander, Jan},
  journal={Industrial Robot: An International Journal},
  volume={32},
  number={1},
  pages={64--70},
  year={2005},
  publisher={Emerald Group Publishing Limited}
}

@article{tactime,
  title={Behavioral biometric optical tactile sensor for instantaneous decoupling of dynamic touch signals in real time},
  author={Son, Changil and Kim, Jinyoung and Kang, Dongwon and Park, Seojoung and Ryu, Chaeyeong and Baek, Dahye and Jeong, Geonyoung and Jeong, Sanggyun and Ahn, Seonghyeon and Lim, Chanoong and others},
  journal={Nature Communications},
  volume={15},
  number={1},
  pages={8003},
  year={2024},
  publisher={Nature Publishing Group UK London}
}

@article{gpt5,
  title={Openai gpt-5 system card},
  author={Singh, Aaditya and Fry, Adam and Perelman, Adam and Tart, Adam and Ganesh, Adi and El-Kishky, Ahmed and McLaughlin, Aidan and Low, Aiden and Ostrow, AJ and Ananthram, Akhila and others},
  journal={arXiv preprint arXiv:2601.03267},
  year={2025}
}

@misc{deepseekv4,
  title  = {DeepSeek-V4: Towards Highly Efficient Million-Token Context Intelligence},
  author = {{DeepSeek-AI}},
  year   = {2026}
}

@misc{ur5,
  author       = {{Universal Robots}},
  title        = {{UR5 Technical Specifications}},
  year         = {2016},
  note         = {Technical specifications, Item no. 110105, EN 09/2016},
  howpublished = {\url{https://www.universal-robots.com/media/50588/ur5_en.pdf}},
  urldate      = {2026-05-24}
}

\appendix

\section{Appendix}
\label{Appendix}

\subsection{Details of TouchThinker-1M Dataset}
\label{Details of TouchThinker-1M Dataset}

\subsubsection{Source Data of TouchThinker-1M}
\label{Source Data of TouchThinker-1M}

\begin{figure*}[!t] 
    \centering 
    \includegraphics[width=\textwidth]{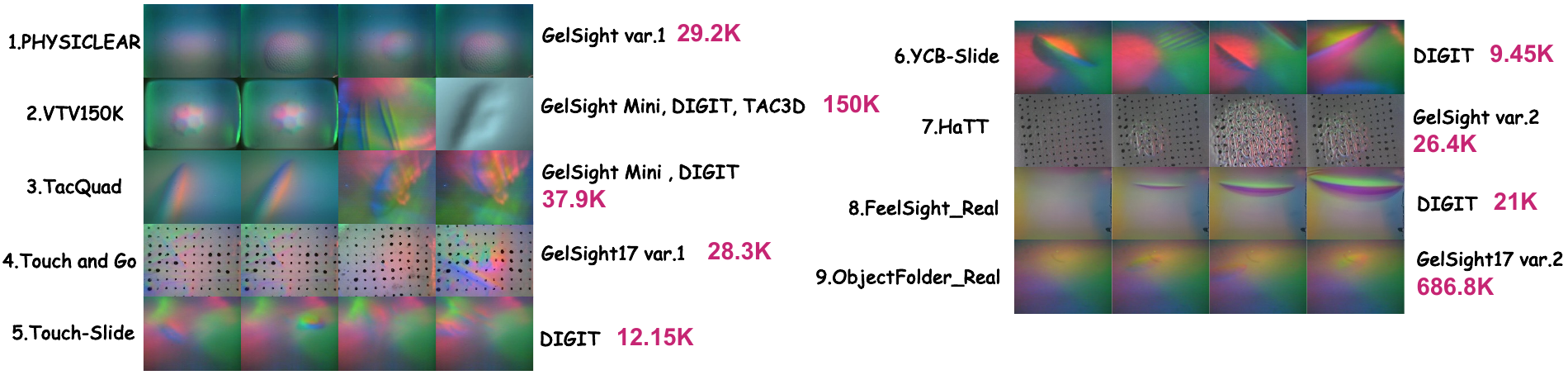}
    \caption{Overview and representative examples of the source datasets in TouchThinker-1M.}
    \label{appendix_1M} 
    \vspace{-3.5mm}
\end{figure*}

TouchThinker-1M integrates nine tactile datasets across seven tactile sensing platforms, four major acquisition actions, and 415 deduplicated objects, comprising 1,001,344 tactile frames in total. Detailed object categories are provided in Figure~\ref{fig:object_touchthinker1m}. We next provide an overview of the source datasets that constitute TouchThinker-1M.

\begin{itemize}[leftmargin=*, itemsep=0pt, parsep=0pt, topsep=2pt, partopsep=0pt]
    \item \textbf{VTV150K}~\citep{vtv}. It contributes 150,000 visuo-tactile video frames from 100 common objects, manually captured with GelSight Mini, DIGIT, and Tac3D sensors through normal pressing, rotational movement, and sliding motion.
    \item \textbf{PhysiCLEAR}~\citep{octopi}. It retains 48 everyday objects after duplicate removal and uses a GelSight var.1 tactile sensor for handheld pressing and rotation, comprising 29,211 tactile frames after curation.
    \item \textbf{Touch and Go}~\citep{tag}. It spans 18 material categories, where human collectors probe real-world objects with a GelSight17 var.1 tactile sensor while recording egocentric video, comprising 28,300 tactile frames after curation.
    \item \textbf{TacQuad}~\citep{anytouch}. It uses the coarse-grained spatially aligned subset covering 92 objects, collected with GelSight Mini and DIGIT image-based tactile sensors through sequential handheld pressing at the same contact location with additional twisting motions, comprising 37,955 tactile frames after curation.
    \item \textbf{Touch-Slide}~\citep{Touch-Slide}. It covers 9 toy-kitchen objects collected as human sliding interaction trajectories with a DIGIT tactile sensor, comprising 12,150 tactile frames after curation.
    \item \textbf{YCB-Slide}~\citep{YCB-Slide}. YCB-Slide covers 7 YCB objects collected by sliding a handheld DIGIT tactile sensor across fixed object surfaces and recording trajectories, comprising 9,450 tactile frames after curation.
    \item \textbf{FeelSight-Real}~\citep{FeelSight}. It covers 5 real-world objects collected with four DIGIT vision-based tactile sensors mounted on the fingertips of an Allegro hand, using an in-hand rotation policy to record vision, touch, and proprioception, comprising 21,000 tactile frames after curation.
    \item \textbf{ObjectFolder-Real}~\citep{objectfolder}. It covers 70 real-world household object instances, collected with GelSight17 var.2 tactile sensors mounted on a Franka Emika Panda arm by contacting selected surface points along their normal directions and recording gel deformation, comprising 686,880 tactile frames after curation.
    \item \textbf{HaTT}~\citep{HaTT}. It covers 66 material textures using GelSight var.2 single-press videos augmented from the original HaTT dataset as tactile data, comprising 26,398 tactile frames after curation.
\end{itemize}

\subsubsection{Tactilc Instruction Data Synthesis}
\label{Tactilc Instruction Data Synthesis}

\noindent \textbf{Template-based instructions.} Following Octopi~\citep{octopi} and VTV-LLM~\citep{vtv}, we construct template-based tactile instructions by instantiating predefined templates with attribute labels and tactile descriptions. The generated data cover basic tactile property understanding and question-answering tasks, including tactile feature analysis (TFA), surface feature distinction (SFD), surface optimality identification (SOI), object sensation correlation (OSC), and tactile scenario analysis (TSA), enabling models to learn tactile-semantic correspondences across surface properties, object sensations, and interaction contexts. However, this template-driven paradigm is constrained by template coverage and expressiveness; its fixed wording and label-based supervision may encourage shallow correlations rather than grounded tactile reasoning.

\noindent \textbf{Tactile Chain-of-Thought Instructions.} Prior work~\citep{octopi,octopi15,vtv} mainly relies on attribute classification or template-based question answering, yielding discrete labels or brief responses without explicitly modeling contact dynamics or causal dependencies. This limitation induces shallow tactile-semantic associations. To address it, we construct a high-quality tactile chain-of-thought reasoning dataset grounded in intrinsic object-state changes during dynamic tactile interactions. The construction pipeline is detailed below. \textbf{Stage 1}: Data Preparation. We first curate tactile videos collected under a unified acquisition protocol, retaining key contact and motion segments that reveal object-property variations. We then standardize task metadata, including question categories, task instructions, and answer annotations, to provide a consistent training format. \textbf{Stage 2}: Chain-of-Thought Generation.
We reformulate tactile understanding as cross-modal question answering and attribute-description generation. Using predefined prompt templates, we guide a large language model~\citep{deepseekv4} to generate reasoning traces grounded in deformation dynamics, contact-region changes, and geometric or texture cues observed in tactile videos. Each output is organized as {\small \texttt{...\textless think\textgreater...\textless/think\textgreater\textless answer\textgreater...\textless/answer\textgreater}}, where the reasoning is aligned with key interaction phases and the answer summarizes the inferred physical property. \textbf{Stage 3}: Manual Filtering. We conduct consistency checking and manual review to ensure data quality. Samples are discarded if their reasoning fails to capture key interaction stages or if the intermediate reasoning conflicts with the final answer, thereby preserving logical consistency between chain-of-thought supervision and attribute prediction. \textbf{Stage 4}: Formatting and Data Integration. Finally, we integrate tactile videos, task annotations, structured prompts, chain-of-thought texts, and final answers into a unified training-ready format. This standardized pipeline enables supervision for dynamic tactile reasoning and improves tactile understanding and decision reliability in open-world scenarios.

\begin{figure*}[!t] 
    \centering 
    \includegraphics[width=\textwidth]{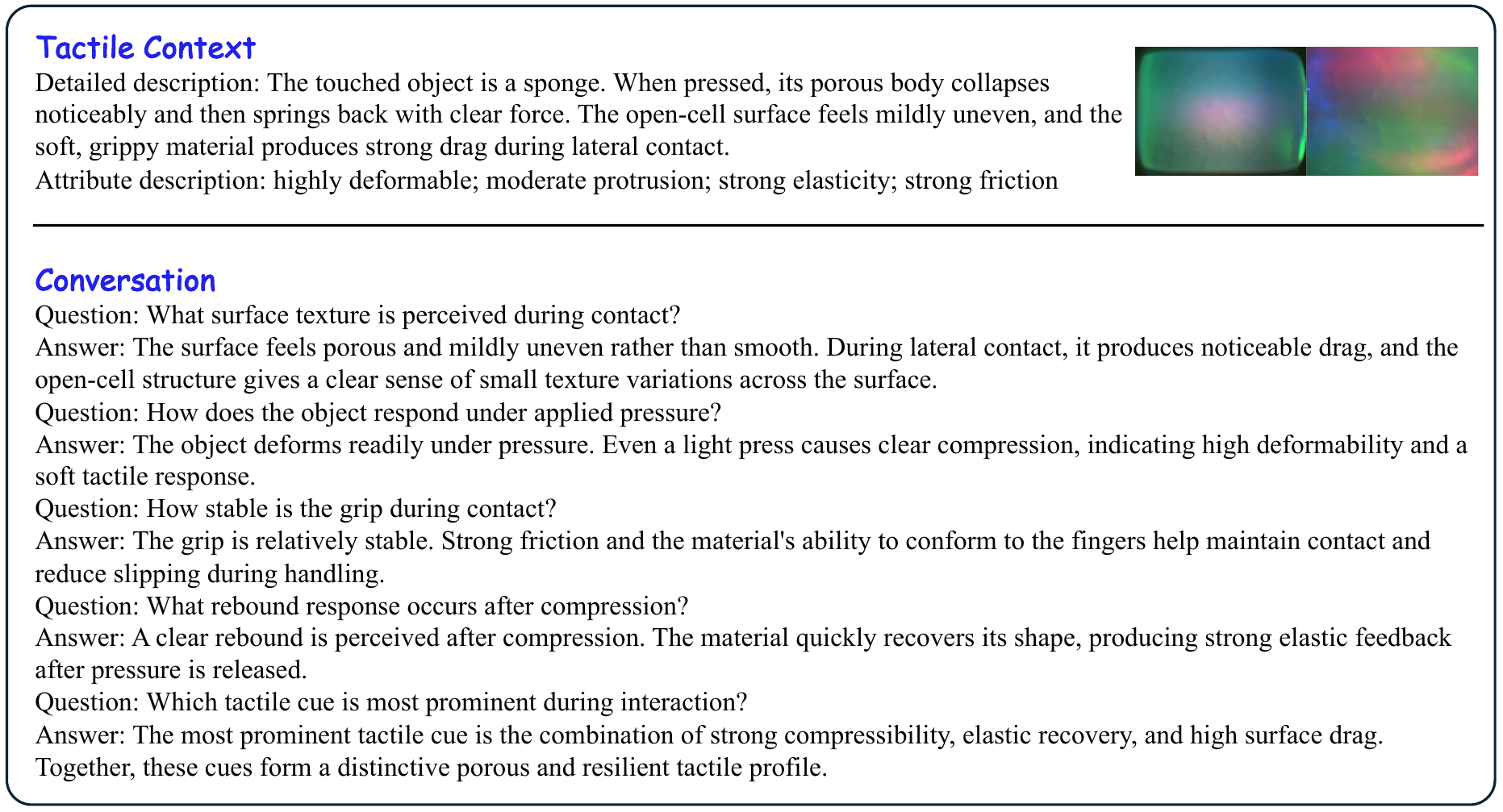} 
    \caption{Example of an open-ended tactile instruction in TouchThinker-1M, using multi-granularity tactile context for grounded reasoning about physical properties and touch-based interactions.}
    \label{appendix_dataset2}
    \vspace{-2mm}
\end{figure*}

\lstdefinestyle{promptstyle}{
  basicstyle=\ttfamily\scriptsize,
  breaklines=true,
  breakatwhitespace=false,
  columns=fullflexible,
  keepspaces=true,
  showstringspaces=false,
  frame=none
}

\begin{figure*}[t]
\centering
\begin{tcolorbox}[
  title={System Prompt},
  width=\textwidth,
  sharp corners,
  colframe=promptgray,
  colback=white,
  colbacktitle=promptgray,
  coltitle=white,
  fonttitle=\bfseries\small,
  boxrule=0.8pt,
  left=6pt,
  right=6pt,
  top=5pt,
  bottom=5pt,
  toptitle=1pt,
  bottomtitle=1pt,
  lefttitle=4pt,
  righttitle=4pt,
  enhanced,
  shadow={2pt}{-2pt}{0pt}{opacity=0.35,promptgray}
]
\begin{lstlisting}[style=promptstyle]
messages = [{"role": "system", "content": f"""You are an AI tactile expert interacting with a single touched object. The input provides the tactile context of this object, which contains two complementary descriptions of the same object: a detailed description and an attribute description. When answering all questions, treat the tactile context as the information perceived during contact.

Design a conversation between you and a person asking about the touched object. The questions should be diverse, and the answers must be grounded only in the given tactile context.

The conversation should include questions about the object's physical tactile properties, such as hardness, protrusion, elasticity, friction, and related aspects. Only ask questions that can be clearly answered from the tactile context or answered reliably.

Also include complex questions related to the touched object, such as questions involving tactile commonsense or interaction and perception during touch. These questions may discuss graspability, pressure feedback, surface discriminability, manipulation stability, and related aspects. Do not ask about uncertain details. Complex answers should be more detailed, clearly structured, and supported by examples or reasoning when necessary."""}]
\end{lstlisting}
\end{tcolorbox}
\caption{Prompt template for generating open-ended tactile instruction data in TouchThinker-1M. The system prompt guides the model to synthesize evidence-grounded, touch-centric dialogues from the supplied tactile context, covering low-level tactile attributes and higher-level interaction-aware perceptual reasoning.}
\label{fig:tactile_instruction_prompt}
\end{figure*}

\noindent \textbf{Open-ended Tactile QA Instructions.} To enhance semantic understanding and interaction-oriented reasoning in open-ended tactile scenarios, we construct tactile instruction data in an open-ended question-answering format, enabling models to learn aligned mappings from tactile context to language responses. Inspired by Cheng et al.~\citep{stola}, we design an automated generation pipeline based on the raw tactile data in TouchThinker-1M. This pipeline builds on object-level tactile data, object identifiers, and manually annotated tactile attribute descriptions, and uses DeepSeek-V4~\citep{deepseekv4} to generate instruction-following samples for tactile interaction. Specifically, we generate open-ended tactile question-answering data from two complementary perspectives: intrinsic tactile attributes, including hardness, protrusion, elasticity, and friction, and interaction-oriented perceptual experiences, including graspability, pressure feedback, surface discriminability, and manipulation stability. This design covers both low-level tactile properties and realistic touch-based reasoning scenarios. We first construct an exemplar pool, where each exemplar contains a tactile context and a grounded dialogue, as illustrated in Figure~\ref{appendix_dataset2}. The tactile context consists of detailed and attribute-level descriptions, and only question-answer pairs explicitly supported by the given tactile evidence are retained. For each target sample, we use its tactile context as the query and retrieve semantically relevant exemplars as few-shot demonstrations. Because TouchThinker-1M organizes tactile samples, object identifiers, and attribute descriptions at the object level, the retrieved demonstrations help align generated responses with the corresponding tactile evidence. The complete prompt construction is shown in Figure~\ref{fig:tactile_instruction_prompt}. We then convert model outputs into single-turn conversations and manually review and correct them, yielding 5,000 touch-language instruction-following samples.

\begin{figure*}[!t] 
    \centering 
    \includegraphics[width=\textwidth]{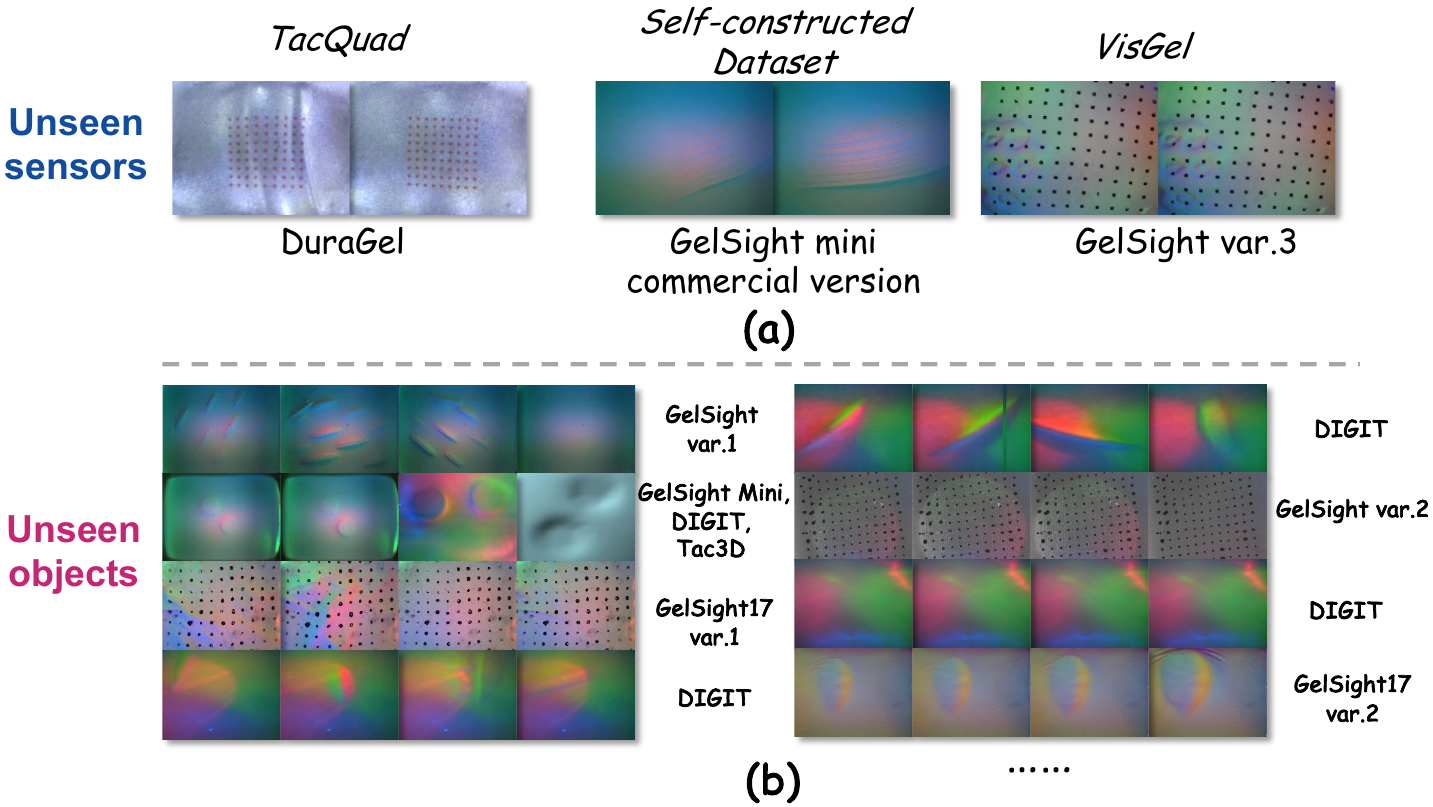} 
    \caption{\textbf{Overview and representative examples of TouchThinker-Bench.} (a) The unseen-sensor split consists of tactile sensors absent from the training set, including DuraGel from TacQuad~\citep{anytouch}, the commercial GelSight Mini from our self-constructed dataset, and GelSight Var. 3 from VisGel~\citep{T3}. These sensors exhibit varying degrees of domain shift from the training sensors in imaging mechanism, color distribution, and gel deformation characteristics, enabling evaluation of cross-sensor generalization. (b) The unseen-object split consists of objects withheld from training, evaluating whether models can generalize to open-world objects with unseen tactile appearances and physical properties.}
    \label{Appendix_touchthinkerbench} 
\end{figure*}

\subsection{Details of TouchThinker-Bench}
\label{Details of TouchThinker-Bench}

\subsubsection{Source Data of TouchThinker-Bench}
\label{Source Data of TouchThinker-Bench}
TouchThinker-Bench consists of two components. The first is derived from the predefined test splits of TouchThinker-1M and contains objects held out from training, enabling evaluation of model generalization to novel objects. The second targets cross-sensor transfer, comprising DuraGel data from the original TacQuad~\citep{anytouch} test split, whose sensor is absent during training, as well as our self-constructed dataset and VisGel~\citep{T3}, collected with the unseen commercial GelSight Mini and GelSight Var. 3 sensors, respectively. Overall, TouchThinker-Bench covers 10 tactile sensors and 82 test objects, spanning both unseen-object and unseen-sensor settings to provide a comprehensive evaluation of tactile perception, attribute reasoning, and cross-sensor generalization. We provide a detailed description of each component below.

\noindent \textbf{Cross-object generalization evaluation.} It consists of multiple existing tactile datasets from the test portion of TouchThinker-1M, as shown in Figure~\ref{Appendix_touchthinkerbench}(b). The PhysiCLEAR~\citep{octopi} portion is collected with a GelSight var.1 tactile sensor and contains 5 daily objects in the test set; the VTV-150K~\citep{vtv} portion is collected with three tactile sensors, including GelSight Mini, DIGIT, and Tac3D, and contains 5 common objects; the Touch and Go~\citep{tag} portion is collected with a GelSight17 var.1 tactile sensor and contains 18 material surfaces; the YCB-Slide~\citep{YCB-Slide} portion is collected with a DIGIT tactile sensor and contains 3 YCB objects; the Touch-Slide~\citep{Touch-Slide} portion is collected with a DIGIT tactile sensor and contains 3 toy-kitchen objects; the HaTT~\citep{HaTT} portion is collected with a GelSight var.2 tactile sensor and contains 9 material textures; the FeelSight-Real~\citep{FeelSight} subset uses four fingertip-mounted DIGIT sensors on an Allegro hand and contains 1 real object, while ObjectFolder-Real~\citep{objectfolder} uses a GelSight17 var.2 sensor and contains 9 real household objects.

\noindent \textbf{Cross-sensor generalization evaluation.} It consists of data collected with three tactile sensors unseen during training, as shown in Figure~\ref{Appendix_touchthinkerbench}(a). The TacQuad subset is collected using the DuraGel tactile sensor and contains 15 test objects~\citep{anytouch}, while the VisGel subset is collected using the unseen GelSight Var. 3 tactile sensor~\citep{T3} and includes 4 objects. To further assess performance in real-world deployment scenarios, we additionally collected tactile data for 10 categories of everyday objects using a UR5 robotic arm~\citep{ur5} equipped with the commercial GelSight Mini sensor.

\subsubsection{Task Types}
\label{Task Types}

\noindent \textbf{Basic Tactile Property Understanding.} This task evaluates whether models can identify fundamental tactile attributes, including hardness, roughness, elasticity, and friction. The evaluation covers both unseen sensors and unseen objects, thereby assessing the generalization ability of tactile representations across sensor domains and object categories.

\noindent \textbf{Basic Tactile Reasoning.} Following prior benchmarks such as Octopi~\citep{octopi} and VTV-LLM~\citep{vtv}, this task includes surface feature distinction (SFD), surface optimality identification (SOI), object sensation correlation (OSC), and tactile scenario analysis (TSA). These tasks enable standardized quantitative comparison of models’ tactile reasoning capabilities.

\noindent \textbf{Open-ended tactile commonsense reasoning.} We construct open-ended tactile question-answering samples from TouchThinker-Bench using tactile images and their category labels. Specifically, we design a unified prompt template for three task types and use category labels and tactile attribute cues as conditional inputs to DeepSeek-V4~\citep{deepseekv4} for generating candidate question-answer pairs, followed by manual verification. The benchmark covers Touch Attribute Understanding, Touch Interaction Understanding, and Touch Knowledge Reasoning. These tasks respectively evaluate models’ ability to recognize basic tactile properties, such as hardness, protrusion, elasticity, and friction; reason about interaction-related characteristics, such as graspability, pressure feedback, surface discriminability, and manipulation stability; and infer higher-level tactile knowledge, including material properties, usage patterns, and everyday interaction scenarios, by integrating tactile cues with object categories. As shown in Figure~\ref{fig:open_qa_prompt}, each question is queried independently to avoid unstable formatting and answer-segmentation errors when generating multiple long-form responses. We then parse, deduplicate, and format the model outputs into single-turn tactile question-answering samples. Finally, we conduct manual review and correction to ensure that each question is supported by tactile and category information and that the reference answers are consistent with the corresponding objects’ tactile attributes and interaction characteristics.


\begin{figure*}[t]
\centering
\begin{tcolorbox}[
  title={Open-ended Tactile QA Prompt},
  width=\textwidth,
  sharp corners,
  colframe=promptblue,
  colback=white,
  colbacktitle=promptblue,
  coltitle=white,
  fonttitle=\bfseries\small,
  boxrule=1.0pt,
  left=6pt,
  right=6pt,
  top=5pt,
  bottom=5pt,
  toptitle=1pt,
  bottomtitle=1pt,
  lefttitle=4pt,
  righttitle=4pt,
  enhanced,
  shadow={2pt}{-2pt}{0pt}{opacity=0.45,promptgray}
]
\scriptsize

\noindent\textcolor{blue}{\textbf{Text Context}}

\noindent
You are an AI tactile assistant interacting with a single touched object. You will understand the touched object from the text prompt, where the object class is \texttt{\textless object class\textgreater}. Imagine that you are physically touching this object.

\vspace{0.6em}

\noindent
Design conversations between you and a person asking about this object. The answers should be written in a tone indicating that you are touching the object and answering based on tactile perception.

\vspace{0.6em}

\noindent
All answers must be based on the confirmed tactile attribute tendencies and the object class, and should not introduce details that cannot be supported by the given information.

\vspace{0.8em}
\hrule
\vspace{0.8em}

\noindent\textcolor{blue}{\textbf{Object Class}}

\noindent
a microfiber cloth

\vspace{0.5em}

\noindent\textcolor{blue}{\textbf{Task Specific Prompt Type 1: Touch Attribute Understanding}}

\noindent
Include questions asking about the typical tactile physical properties of the touched object, including hardness, protrusion, elasticity, and friction. Design a separate conversation for each tactile attribute. Ensure that the conversations contain only confirmed information.

\vspace{0.3em}

\noindent
Required format:

\begin{quote}
\ttfamily\scriptsize
[\\
\quad \{"question": "\textless question 1\textgreater", "answer": "\textless answer\textgreater"\},\\
\quad \{"question": "\textless question 2\textgreater", "answer": "\textless answer\textgreater"\},\\
\quad \{"question": "\textless question 3\textgreater", "answer": "\textless answer\textgreater"\},\\
\quad \{"question": "\textless question 4\textgreater", "answer": "\textless answer\textgreater"\},\\
\quad \{"question": "\textless question 5\textgreater", "answer": "\textless answer\textgreater"\}\\
]
\end{quote}

\vspace{0.3em}

\noindent\textcolor{blue}{\textbf{Task Specific Prompt Type 2: Touch Interaction Understanding}}

\noindent
Include questions asking about interaction and perception characteristics during touch, including graspability, pressure feedback, surface discriminability, and manipulation stability. Randomly select three of these interaction and perception characteristics, and design a separate conversation for each. Each conversation should include one question and 3--5 corresponding ground-truth answers. Ensure that the conversations contain only confirmed information.

\vspace{0.3em}

\noindent
Required format:

\begin{quote}
\ttfamily\scriptsize
[\\
\quad \{"question": "\textless question 1\textgreater", "answer": "\textless answer\textgreater"\},\\
\quad \{"question": "\textless question 2\textgreater", "answer": "\textless answer\textgreater"\},\\
\quad \{"question": "\textless question 3\textgreater", "answer": "\textless answer\textgreater"\}\\
]
\end{quote}

\vspace{0.3em}

\noindent\textcolor{blue}{\textbf{Task Specific Prompt Type 3: Touch Knowledge Reasoning}}

\noindent
Include complex reasoning questions involving background knowledge related to the touched object, emphasizing the tactile perspective. Design two conversations. Each conversation should include one question and 3--5 candidate ground-truth answers, with each answer being independent and self-contained. Ensure that the conversations contain only confirmed information. Provide more detailed answers for complex questions. For example, include concrete examples or reasoning steps to make the content more convincing and well-organized. Multiple paragraphs may be used if necessary.

\vspace{0.3em}

\noindent
Required format:

\begin{quote}
\ttfamily\scriptsize
[\\
\quad \{"question": "\textless question 1\textgreater", "answer": "\textless answer\textgreater"\},\\
\quad \{"question": "\textless question 2\textgreater", "answer": "\textless answer\textgreater"\}\\
]
\end{quote}

\end{tcolorbox}
\caption{Prompt template for open-ended question answering in TouchThinker-Bench. The upper section presents the multimodal input to DeepSeek-V4~\citep{deepseekv4}, including a textual prompt and its paired visual input.}
\vspace{-2mm}
\label{fig:open_qa_prompt}
\end{figure*}

\begin{table}[!t]
\centering
\small
\begin{tabular}{lcc}
\toprule
\textbf{Configuration} & \textbf{Stage I} & \textbf{Stage II} \\
\midrule
Optimizer & AdamW  & AdamW \\
Learning Rate & 2e-4 & 2e-4 \\
Weight Decay & 0.001 & 0.001 \\
Training Epochs & 1 & 1 \\
Warmup Ratio & 0.1 & 0.1 \\
Learning Rate Scheduler & Linear & Linear \\
Batch Size Per GPU & 16 & 16 \\
Maximum Token Length & 512 & 512 \\
Unfreeze LLM & \ding{55} & \checkmark \\
\bottomrule
\end{tabular}
\caption{Training configurations for the two-stage training paradigm. Stage I: Tactile-Text Alignment. Stage II: End-to-end Supervised Fine-Tuning.}
\label{tab:training_config}
\end{table}

\subsection{Details of Implementation}
\label{Details of Implementation}

\subsubsection{Configuration and Hyperparameters}
\label{Configuration and Hyperparameters}

We present the detailed training configuration and hyperparameters in the two stages in Table~\ref{tab:training_config}.

\subsubsection{GPT-5 and DeepSeek-V4 scoring}
\label{GPT-5 and DeepSeek-V4 scoring}
Since TouchThinker-Bench is designed for open-ended and free-form tactile commonsense reasoning, model responses may vary substantially in wording, level of detail, and reasoning style. Therefore, conventional n-gram-based metrics such as CIDEr and BLEU-4 are insufficient to fully capture the quality of generated answers. To provide a more comprehensive evaluation, we report METEOR as a semantic-similarity-oriented metric and further employ GPT-5~\citep{gpt5} and DeepSeek-V4~\citep{deepseekv4} as automatic judges for multidimensional response assessment.

Specifically, each generated response is evaluated from five perspectives: semantic correctness, tactile consistency, commonsense and reasoning plausibility, information completeness, and language quality. For each dimension, the judge assigns a score from 1 to 10 and provides a brief rationale. The final score is computed as the average of the five dimension-level scores:
\begin{equation}
    S = \frac{1}{5}\sum_{i=1}^{5} s_i,
\end{equation}
where $s_i$ denotes the score of the $i$-th evaluation dimension. The evaluation dimensions are defined as follows:

\begin{itemize}[leftmargin=*, itemsep=0pt, parsep=0pt, topsep=2pt, partopsep=0pt]
    \item \textbf{Semantic Correctness:} Whether the response is semantically consistent with the reference answer and free from factual errors.
    \item \textbf{Tactile Consistency:} Whether the response accurately reflects tactile perception, physical properties, or interaction-related sensations, and whether it is consistent with the provided tactile information.
    \item \textbf{Commonsense and Reasoning Plausibility:} Whether the response accords with real-world physical commonsense and presents a reasonable and convincing reasoning process.
    \item \textbf{Information Completeness:} Whether the response covers the key information required by the question and avoids omitting important tactile attributes or explanations.
    \item \textbf{Language Quality:} Whether the response is clear, natural, and fluent, without obvious grammatical or expression errors.
\end{itemize}

\subsection{Additional Ablation Studies}
\label{Additional Ablation Studies}

\noindent \textbf{Impact of the model scale.} To investigate the impact of model scale on visuo-tactile understanding, we evaluate LLM backbones of varying sizes and report the performance of TouchThinker with Qwen-2.5-14B, as shown in Table~\ref{vtv_task}. The results demonstrate consistent improvements across most subtasks as model size increases, with especially significant improvements on reasoning-oriented tasks such as SFD.

\noindent \textbf{Impact of Question-Guided Token Fusion and Gaussian Temporal MoE in the Action-Aware Modeling Mechanism.} We conduct ablation experiments on VTV-150K to validate the effectiveness of the two core components in our action-aware modeling mechanism. As shown in Table~\ref{tab:ablation_action_aware}, both components contribute to tactile reasoning performance. Removing Question-Guided Token Fusion prevents the model from incorporating task semantics during tactile feature aggregation, leading to less focused temporal representations and degraded reasoning accuracy. Removing Gaussian Temporal MoE also leads to performance degradation, indicating that uniform temporal modeling is insufficient for capturing action-specific tactile evidence. In contrast, combining Question-Guided Token Fusion with Gaussian Temporal MoE achieves the best overall performance. This demonstrates that early question-aware fusion helps suppress irrelevant tactile segments, while Gaussian Temporal MoE further localizes task-relevant action intervals, enabling more efficient tactile representations and stronger downstream reasoning.

\begin{table}[!t]
\centering
\small
\setlength{\tabcolsep}{3pt}
\renewcommand{\arraystretch}{1.08}
\resizebox{\columnwidth}{!}{
\begin{tabular}{cc|ccccc}
\toprule
\textbf{Question-Guided} & \textbf{Gaussian} 
& \textbf{SFD} & \textbf{SOI} & \textbf{OSC} & \textbf{TSA} & \textbf{Avg.} \\
\textbf{Token Fusion} & \textbf{Temporal MoE} 
&  &  &  &  &  \\
\midrule
\checkmark & \checkmark 
& \textbf{78.9} & \textbf{64.2} & \textbf{50.7} & \textbf{74.0} & \textbf{67.0} \\
            & \checkmark 
& 73.4 & 51.6 & 42.2 & 70.0 & 59.1 \\
\checkmark &             
& 71.2 & 52.9 & 45.8 & 72.0 & 60.5 \\
\bottomrule
\end{tabular}
}
\caption{Ablation study of Question-Guided Token Fusion and Gaussian Temporal MoE in the action-aware modeling mechanism.}
\label{tab:ablation_action_aware}
\end{table}

\begin{figure*}[t]
\centering
\begin{tcolorbox}[
  title={Object Categories in TouchThinker-1M},
  width=\textwidth,
  sharp corners,
  colframe=black!70,
  colback=white,
  colbacktitle=black!70,
  coltitle=white,
  fonttitle=\bfseries\small,
  boxrule=0.8pt,
  left=6pt,
  right=6pt,
  top=5pt,
  bottom=5pt,
  enhanced
]
\scriptsize
\noindent
tangerine, carton, playing card, orange, spoon with rice, nectarine, tomato, rubber glove, pen pad, cotton cloth, velvet, sandpaper, chip bag, rubber slipper sole, candle, suede, silk scarf, fascia ball, chalk, clay, avocado, lemon, banana, kiwi, pineapple, plastic bottle, waffle, sponge, plastic basket, balloon, leather glove, building block protrusion, jelly, piano key, blanket, ceramic cup, oven glove, bark, scouring pad, fur, pine cone, ping pong ball, plastic building block, cork, wooden ruler, eraser, velcro, leather wallet, toilet paper roll, shower mat, baseball, golf ball, sticky note, silicone pad, yoga mat, masking tape roll, rubber band, cotton ball, gauze, computer mouse, headphone, face towel, woven watch strap, rubber watch strap, metal watch strap, wooden block, marble, claw, keyboard, remote control button, toothbrush head, vitamin tablet, tennis ball, towel, absorbent cloth, wrist guard, fine bubble film, coarse bubble film, rice, ridge cup, key, screw, circuit board, mold, sponge sheet, wrench, screwdriver handle, aluminum tube, hairpin, steel wool, huamei, candy, grid bag, ridged plastic bottle, scissors, iron ruler, iron clip, tape, injection tube, wire, scissors handle, dishwashing cloth, mesh strainer scoop, toothbrush handle, scissors blade, TV remote back, pillow, disposable water bottle, game controller thumbstick, tissue paper, mop head, rice spatula handle, game controller buttons, hairbrush bristles, leather book cover, bread knife handle, card holder, stress ball, feather duster head, toilet brush handle, potato, paper towel, game controller shoulder buttons, lanyard, insulating holder, basket, rubber slippers, rice spatula scoop, aluminium foil, pen barrel, mesh strainer handle, egg, game controller keypad, bread knife blade, game controller body, bubble wrap, ice block, clothes peg, TSA lock numbers, millet, nylon shirt, feather duster handle, jeans, microfiber cloth, hairbrush handle, toilet brush bristles, hairbrush bristle ends, bath towel, concrete surface, plastic surface, glass surface, wooden surface, metal surface, brick surface, tile surface, leather surface, synthetic fabric surface, rubber surface, paper surface, tree bark surface, grass surface, soil surface, rock surface, gravel surface, sand surface, plant surface, 3D-printed object, USB connector, vacuum cup, vacuum cup lid, glass jar, bowl edge, bowl handle, metal can, metal can bottom, metal can top, CD, coffee cup bottom edge, cup shell edge, cup shell interior, glass bottle, hot dog, red metal box, metal box edge, plush toy, metal box, mouse scroll wheel, onion, paper cup, peach, pencil box, pill bottle bottom edge, pill bottle cap, plastic bottle cap, plastic box bottom, duck toy beak, duck toy wing, rubber rabbit toy, rubber rabbit ear, rubber ball, shampoo bottle, glove, sponge brush handle, sponge brush, spoon, plastic box, plastic cup, plastic glove, tomato leaf, wooden cup, rubber toy, shuttlecock feather, marble slab, bamboo surface, bicycle basket, bicycle handle, bicycle pedal, bicycle seat, bicycle wheel, stone, block-patterned carpet, board brush, cable path, chair, desk, door handle, file envelope, glass door, hinge, bamboo leaf, metal post, metal sign, metal block, paper board, plastic chair, post box, tree branch, stone surface, stone pier, stopcock, tube, van surface, wall corner, well cap, wooden stair surface, wall surface, column, bowl, bowl bottom, coffee cup bottom, grape, paper bag, plastic box bottom edge, fiber chair, lock, metal net, metal stair surface, roadblock, toy banana, toy cheese, toy corn, toy lettuce, toy strawberry, toy tomato, toy bread, toy cookie, toy plum, mustard bottle, cleanser bottle, mug, power drill, hammer, sugar box, tomato soup can, salad plate, dinner plate, hair comb, decorative plate, soup ladle, salad fork, frying spatula, 8-inch skillet, 10.25-inch skillet, 10.5-inch griddle, dutch oven, dutch oven lid, rinsing cup, hand scoop, shovel toy, round plate, square plate, cutting board, wine glass, drinking cup, portion cup, cake pan, loaf pan, pestle, mortar, sculpture, ladle, spatula, decorative cast, small fork, large fork, soap dish, beer glass, large container, medium container, small container, vase, plate handle, plate, plate base, display stand, drop funnel, container lid, food pan, large flowerpot, small flowerpot, green vase, blue vase, orange vase, large swan, small swan, spoon holder, utensil container, potato masher, skimmer, pasta server, solid turner, slotted turner, green glass, red glass, scoop, box lid, Stanford frisbee, kettlebell, trim removal tool, bell pepper, dice, pear, pepper grinder, Rubik's cube, ABS plastic, aluminum, artificial grass, athletic shirt fabric, balsa wood, binder, brick, bubble envelope, CD sleeve, canvas, carbon fiber, ceramic, acrylic, coffee filter, cotton, dot paper, EPDM foam, felt, file portfolio, flannel, fleece, floor tile, wooden floor, folder, frosted acrylic, gift box, glitter paper, greeting card, leather back, leather front, MDF, metal foil, metal mesh, metal shelving, foam, wood, nylon bag, nylon mesh, nylon strap, package foam, painted brick, painted wood, green painted wood, pink foam, plastic mesh, pleather, polyethylene foam, portfolio cover, resin carbon fiber, resume paper, rough paper plate, silk, smooth paper plate, stained wood, stone tile, styrofoam, tarp, terra cotta, textured cloth, textured metal, textured paper, textured rubber, tissue paper, vinyl, wax paper, whiteboard.
\end{tcolorbox}
\caption{Object categories in TouchThinker-1M. The figure lists object categories after removing articles, dataset sources, and non-essential descriptive modifiers for compact presentation.}
\label{fig:object_touchthinker1m}
\vspace{-2mm}
\end{figure*}

\end{document}